\documentclass{article}

\usepackage{microtype}
\usepackage{graphicx}
\usepackage{booktabs} 
\usepackage{breqn} 
\usepackage{amsmath} 
\usepackage{changepage}
\usepackage{subfig}

\usepackage{hyperref}
\usepackage{subfig}
\usepackage[T1]{fontenc}

\def\one{\mbox{1\hspace{-4.25pt}\fontsize{12}{14.4}\selectfont\textrm{1}}} 

\usepackage[accepted]{icml2024}

\usepackage{amsmath}
\usepackage{amssymb}
\usepackage{mathtools}
\usepackage{amsthm}
\usepackage{comment}

\usepackage[capitalize,noabbrev]{cleveref}

\theoremstyle{plain}
\newtheorem{theorem}{Theorem}[section]

\newtheorem{lemma}[theorem]{Lemma}

\theoremstyle{definition}
\newtheorem{definition}[theorem]{Definition}

\theoremstyle{remark}

\usepackage[textsize=tiny]{todonotes}

\icmltitlerunning{Bridging Contrastive Learning and Domain Adaptation: Theoretical Perspective and Practical Application}

\begin{document}

\twocolumn[
\icmltitle{Bridging Contrastive Learning and Domain Adaptation: Theoretical Perspective and Practical Application}




\begin{icmlauthorlist}
\icmlauthor{Gonzalo Iñaki Quintana}{labo,comp}
\icmlauthor{Laurence Vancamberg}{comp}
\icmlauthor{Vincent Jugnon}{comp}
\icmlauthor{Agnès Desolneux}{labo}
\icmlauthor{Mathilde Mougeot}{labo,sch}
\end{icmlauthorlist}

\icmlaffiliation{labo}{Centre Borelli, CNRS \& ENS Paris-Saclay, F-91190 Gif-sur-Yvette, France}
\icmlaffiliation{comp}{GE HealthCare, 78530 Buc, France}
\icmlaffiliation{sch}{ENSIIE, 91000 Evry, France}

\icmlcorrespondingauthor{Gonzalo Iñaki Quintana}{gonzalo.quintana@ens-paris-saclay.fr}
\icmlkeywords{Machine Learning, ICML}

\vskip 0.3in
]



\printAffiliationsAndNotice{}  

\begin{abstract}
This work studies the relationship between Contrastive Learning and Domain Adaptation from a theoretical perspective. The two standard contrastive losses, NT-Xent loss (Self-supervised) and Supervised Contrastive loss, are related to the Class-wise Mean Maximum Discrepancy (CMMD), a dissimilarity measure widely used for Domain Adaptation. Our work shows that minimizing the contrastive losses decreases the CMMD and simultaneously improves class-separability, laying the theoretical groundwork for the use of Contrastive Learning in the context of Domain Adaptation. Due to the relevance of Domain Adaptation in medical imaging, we focused the experiments on mammography images. Extensive experiments on three mammography datasets - synthetic patches, clinical (real) patches, and clinical (real) images - show improved Domain Adaptation, class-separability, and classification performance, when minimizing the Supervised Contrastive loss.
\end{abstract}

\section{Introduction}

Given a source data distribution or domain, we are often interested in transferring the representation learned to a different, albeit related, target domain. This is crucial for leveraging models pre-trained on large annotated datasets, as well as for adapting test and training distributions, which are generally different \citep{DAframework}. In particular, Domain Adaptation (DA) methods seek to minimize the effects of the domain shift to enable more efficient transfer. 
This is especially relevant in the medical imaging domain, where high data variability and limited access to large datasets pose significant challenges to the development of Deep Learning (DL)-based solutions, often hindering model generalization and performance across diverse clinical settings \cite{GarruchoDomainGeneralization2022}.

Contrastive Learning (CL) is a learning paradigm where semantically similar data-points are close to one another in the feature space, enabling to learn representations that are invariant given certain transformations. Intuitively, mapping data points from different domains to the same region in the feature space mirrors the DA problem. In addition CL separates the representations of semantically different data-points, which has been found to be beneficial for downstream task performance, like classification, detection, or segmentation. Contrastive Learning has been widely applied in the medical imaging domain \cite{ChaitanyaContrastiveLerMedicalImaging, MICCAI2021_fed_3, CaoMICCAI2021, Quintana2024}.

Inspired by the similarity of the tasks that Domain Adaptation and Contrastive Learning pursue, as well as by the growing interest in CL for DA, we analyze both paradigms to provide theoretical justifications for applying CL to DA. Due the relevance of Domain Adaptation in medical imaging, we conduct experiments on mammography images for classification tasks, specifically determining the presence or absence of breast cancer.

\subsection{Related work}

\paragraph{Domain Adaptation.} Let $\mathcal{D}_s=\{ \mathcal{X}_s \times \mathcal{Y}_s, \pi_s \}$ be a source domain and $\mathcal{D}_t=\{ \mathcal{X}_t \times \mathcal{Y}_t, \pi_t \}$ a target domain, where $ \mathcal{X}_. $ is an instance or covariate space, $ \mathcal{Y}_. $ is the label space, and $ \pi_.: \mathcal{X}_. \times \mathcal{Y}_. \rightarrow \mathbb{R}$ a joint probability measure.The target domain $ \mathcal{D}_t $ is typically unlabeled, i.e., $\mathcal{Y}_t = \emptyset$, contains fewer labels than $\mathcal{D}_s$, or has a smaller dataset. In Domain Adaptation, we seek to transfer the representations learned in the source domain for solving a source task $ \mathcal{T}_s $ to the target domain, while considering that source and target tasks are the same. Various DA strategies have been proposed based on the nature of the domain shift (e.g., covariate shift, prior probability shift, concept shift), the availability of labels in the target domain (supervised, unsupervised, semi-supervised), and the type of models employed (e.g., shallow or deep architectures). In this work we consider the \textit{hidden covariate shift} \citep{deMathelin2023book} or \textit{covariate observation shift} \cite{Kull2014PatternsOD}, a subtype of concept shift where it is assumed that there exists a non-linear transformation of the covariates that eliminates the shift. One of the most widely used DA method consists in aligning domains by minimizing a domain dissimilarity measure, such as the Mean Maximum Discrepancy (MMD) \citep{Gretton2006}, the Kullback-Leibler divergence, the Wasserstein distance \cite{wasserstein_domain_adapt, wasserstein_domain_adapt_2}, or the Bregman divergence \cite{farahani2021brief}. \citet{long2013transfer} first proposed Deep Adaptation Network (DAN), where the MMD is used to minimize the marginal distribution shift, and extended it to matching both the marginal and conditional distributions with Joint Adaptation Network (JAN) \cite{long2017deep}. However, minimizing these dissimilarity measures has been reported to attain DA at the expense of reducing feature-label correlation, decreasing class-separability in the feature space and negatively impacting downstream tasks like classification or detection \cite{wang2021unified}. Domain-adversarial Neural Network (DANN) \citep{ganin2016domainadversarial} and its variants \citep{CDANN, shen2018wassersteinDANN, tzeng_ADDA} consists in jointly training an encoder, classifier, and domain discriminator to obtain domain invariant representations. These adversarial methods require training an additional network with an unstable min-max loss, which often demands extra training time and computational resources \citep{Kouw_2021}.

\paragraph{Contrastive Learning}consists in learning representations where positive pairs of features are dragged to the same region of the feature space, and negative pairs are pushed apart. The definition of positive and negative pairs differs on each application and on the availability of annotations. CL was first introduced as a max-margin loss for dimensionality reduction \citep{contrastive_loss_1, chopra_2005}, which later evolved into the triplet \citep{Weinberger_triplet_loss} and N-pair-mc \citep{n_pair_mc_loss}, which improved convergence and removed the need for negative hard mining. The NT-Xent loss \citep{NT-Xent_loss}, a temperature-scaled version of the N-pair-mc loss, is currently one of the most widely used losses for Self-supervised representation learning, both in Computer Vision \cite{NT-Xent_loss, infoNCE} and in Natural Language Processing \cite{gao2021simcse}. In this context, positive pairs are typically transformed version of the same instance, while negative pairs are all pairs of features that originate from different instances. It has also been used for multimodal learning with text-image aligned representations, which has applications in zero-shot image classification \citep{CLIP, jia2021scaling}, image retrieval \cite{huang2024llm2clip, schall2024optimizing}, and text-conditioned image generation \cite{rombach2022high}. More recently, the Supervised Contrastive loss was introduced \cite{sup_contr_ler} to enable features from different instances to be mapped closely in the feature space. In this case, positive pairs come from instances with the same label, and negative pairs from instances with different labels \cite{sup_contr_ler, li2022selective}. Recently, Contrastive Learning has started gaining traction as a Domain Adaptation method \citep{Thota_2021_CVPR, darban2024dacad, singh2021clda}. However, despite promising empirical results, a theoretical understanding of the DA capabilities of Contrastive Learning is lacking.

\paragraph{Mammography image classification} is crucial for improving screening or diagnostic workflow and accuracy. Its clinical applications span from triaging normal (lesion-free) and abnormal exams to enhance image reading efficiency, to assessing the likelihood of breast cancer and recommending biopsy procedures \cite{kyono2020improving}
Today's state-of-the-art models rely on Convolutional Neural Network (CNN) patch-based approaches: a Deep Learning model is first pre-trained for patch classification and then extended to full-image classification by adding additional layers and re-training \cite{Shen_2019, petrini_2021, Quintana2023}. Currently, there is no publicly available reference dataset of digital mammograms, primarily due to the high cost of obtaining sufficiently large and diverse annotated datasets. DL models typically achieve an AUC ranging from 0.75 to 0.81 for benign vs. malignant classification \cite{bobowicz2023attention, petrini_2021, Shen_2019}. Multi-view models increase that range to 0.83-0.89 by leveraging different views of the same breast and bi-lateral asymmetries between left and right breasts, and using ensembling \cite{wu_2019, bobowicz2023attention, petrini_2021}. In this work, we focus on studying CL and DA and not on establishing a new benchmark performance.

\subsection{Main contributions}

The main contributions of this work are the following:

\begin{itemize}
    \item We show that minimizing the NT-Xent loss and the Supervised Contrastive loss decreases the CMMD, thus improving Domain Adaptation.
    \item We show that minimizing the contrastive losses improves class-separability, by extending the work of \citet{li2021selfsupervised}.
    \item We validate these theoretical results by conducting experiments in a concrete mammography image classification application, using three distinct datasets and the Supervised Contrastive loss.
    \item We introduce a synthetic mammography image dataset based on Gaussian textures and simple lesion simulation. The dataset, along with the code for its generation and for reproducing the experiments in this work, can be found in this repository: \href{github.com/gonzaq94/contrastive-da-synthetic-patch}{github.com/gonzaq94/contrastive-da-synthetic-patch}.
\end{itemize}

\section{Contrastive Learning and dissimilarity measures}
\label{sec:contr_ler_diss_measures}

Consider a learning problem with data from two labeled domains $\mathcal{D}_0 = \{ \mathcal{X} \times \mathcal{Y}, \pi_0 \}$ and $\mathcal{D}_1 = \{ \mathcal{X} \times \mathcal{Y}, \pi_1 \}$, with $ \pi_d: \mathcal{X} \times \mathcal{Y} \rightarrow [0, 1]$ the joint probability measure of the instances $x \in \mathcal{X} \subseteq \mathcal{R}^{n_1 \times n_2} $ and labels $y \in \mathcal{Y}$ of the $d$-th domain. We denote by $ \pi^{\mathcal{X}}_d$ and $ \pi^{\mathcal{Y}}_d$ the marginal probability measures on the instances and labels, and by $ \pi^{\mathcal{X} | \mathcal{Y}}_{d,c}$, $c \in \mathcal{Y}$, the conditional probability measure on the instances knowing the label is $c$. We also consider the mixture domain $ \mathcal{D}_p = \{ \mathcal{X} \times \mathcal{Y}, \pi_p \} $ with joint probability measure $\pi_p \coloneqq p \pi_1  + (1-p) \pi_0$, where $p$ is the mixture probability. In this work, we mostly consider the equiprobable domains case $p=0.5$.

Let $\phi: \mathcal{X} \rightarrow \mathcal{Z} = \mathbb{R}^{m} $ be a feature map parametrized by a neural network, where $m$ is the embedding dimension. $\phi$ defines a Reproducing Kernel Hilbert Space (RKHS) wih kernel $k$ such that $k(x, x') = \langle \phi(x) \; , \; \phi(x') \rangle_{\mathcal{Z}}$, where $\langle . , . \rangle_{\mathcal{Z}} $ denotes the inner product in $ {\mathcal{Z}} $.

\subsection{Contrastive Learning}

We recall the definitions of the Normalized Temperature-scaled Cross Entropy (NT-Xent) loss and the Supervised Contrastive loss.

\begin{definition}[NT-Xent loss]
\label{def:self_sup_contr}

Consider a batch of instances $ \mathcal{B} $ and their feature representation $z$, which are assumed of unitary norm, i.e., $ \| z \| = 1 $ . The NT-Xent loss defined as:
\begin{equation}
    \label{eq:def_self_sup_contr}
    \mathcal{L}_{NT-Xent} = - \frac{1}{| \mathcal{B} |} \sum_{i = 0}^{| \mathcal{B} | - 1} \log \frac{e^{z_i \cdot z_{j(i)} / \tau}}{\sum_{l \in \mathcal{A}(i)} e^{z_i \cdot z_l / \tau}},
\end{equation}
where $ z_i = \phi (x_i) $ is the feature representation of instance $ x_i $, $ z_{j (i)} $ is the positive counterpart of feature $ z_i $, and $ \mathcal{A}(i) = \{ 0, ... | \mathcal{B}| -1 \} \setminus \{ i \} $ is the set of the indices of all features with the exception of $z_i$, and $\tau$ is a temperature parameter. 

\end{definition}

\begin{definition}[Supervised Contrastive loss]
\label{def:sup_contr}

Given a batch of instances $ \mathcal{B} $, the Supervised Contrastive loss is defined as:
\begin{equation}
    \label{eq:def_sup_contr}
    \mathcal{L}_{SupContr} = - \frac{1}{| \mathcal{B} |} \sum_{i \in | \mathcal{B} | } \frac{1}{| \mathcal{P}(i) |} \sum_{j \in \mathcal{P}(i)} \log \frac{e^{z_i \cdot z_j / \tau}}{\sum_{l \in \mathcal{A}(i)} e^{z_i \cdot z_l / \tau}},
\end{equation}
where $ z_i = \phi (x_i) $ is the feature representation of instance $ x_i $, and $ \mathcal{P}(i) = \{ j \in \mathcal{A}(i) : y_j = y_i \} $ is the set of indices of the positive counterparts of feature $ z_i $.

\end{definition}

\subsection{Contrastive Learning and Domain Adaptation}

In the following, we revisit the definition of the CMMD and establish its connection to contrastive losses.

\begin{definition}[CMMD]
\label{def:cmmd}

Given two labeled domains $\mathcal{D}_0$ and $\mathcal{D}_1$, and the mapping $\phi: \mathcal{X} \rightarrow \mathcal{Z}$. The CMMD is defined as:
\begin{dmath}
    \label{eq:cmmd_def}
    \text{CMMD}^2(\mathcal{D}_0, \mathcal{D}_1, \phi) = \mathbb{E}_{C \sim \pi^{\mathcal{Y}}} \left[ \left\| \mathbb{E}_{X \sim \pi^{\mathcal{X}|\mathcal{Y}}_{0,C}} [ \phi(X) ] - \mathbb{E}_{X \sim \pi^{\mathcal{X}|\mathcal{Y}}_{1,C}} [  \phi(X) ] \right\|_{\mathcal{Z}}^2 \right].
\end{dmath}
    
\end{definition}

The CMMD calculates the difference between the expected embedding of instances in the two domains, for each class. If $ \phi $ is adjusted so as to minimize the CMDD, then the embeddings $ \phi (X) $ are similar regardless of the domain, and the conditional distributions of the embeddings of the two domains will be matched. It can thus be seen as a measure of Domain Adaptation. Definition \ref{def:cmmd} corresponds to the definition of the Weighted Class-wise MMD (WCMMD) and encompasses other definitions of the CMMD found in the literature \citep{wang2021unified} as a particular case when all the classes have the same prior probability. We propose the following lemma that relates Contrastive Learning to the minimization of the CMMD (proof in Appendix \ref{app:proof-lemma-contr-cmmd}).

\begin{lemma}
    
\label{lemma:sup_contr_and_cmmd}

In a high temperature regime, both the Supervised Contrastive loss and the NT-Xent loss can be expressed in terms of the CMMD by the following equation:
{
\begin{dmath}
\label{eq:sup_contr_and_cmmd}
    \tau \mathcal{L}_{Contr} 
    \approx
    \frac{1}{4} \text{CMMD}^2(\mathcal{D}_0, \mathcal{D}_1, \phi) +\underbrace{\mathbb{E}_{X, X'  \sim \pi^{\mathcal{X}}_{0.5}} \left[ k(X, X') \right]}_{A} - \frac{1}{2} \underbrace{\mathbb{E}_{C \sim \pi^{\mathcal{Y}}} \left[ \mathbb{E}_{X,X' \sim \pi^{\mathcal{X}|\mathcal{Y}}_{0, C}} \left[ k(X,X') \right] + \mathbb{E}_{X,X' \sim \pi^{\mathcal{X}|\mathcal{Y}}_{1, C}} \left[ k(X,X') \right] \right]}_{B} + \frac{1}{2\tau} \underbrace{\mathbb{E}_{X \sim \pi^{\mathcal{X}}_{0.5}} \left[ \mathrm{Var}_{X'  \sim \pi^{\mathcal{X}}_{0.5}}  \left[ k(X, X') \right] \right]}_{C} + \mathcal{O} \left( \frac{\mathbb{E}_{X  \sim \pi_{0.5}} \left[ {\mathrm{Var}}_{X'  \sim \pi^{\mathcal{X}}_{0.5}} \left[ k(X, X') \right]^2 \right]}{\tau^4} \right) + \log \left( | \mathcal{B} | -1 \right).
\end{dmath}
}
\end{lemma}

Lemma \ref{lemma:sup_contr_and_cmmd} suggests that decreasing $\mathcal{L}_{contr}$ decreases the CMMD, which improves Domain Adaptation. Equation \eqref{eq:sup_contr_and_cmmd} also includes other terms, which can be analyzed as follows: $A$ represents the similarity between all pairs of features, while $B$ denotes the similarity between pairs of features from the same class and domain. $C$ is a variance term. The constant term $\log \left( | \mathcal{B} | -1 \right)$, with $ | \mathcal{B} | $ and $ \tau $ the batch size and the temperature of the Contrastive losses, is irrelevant for the optimization. The last term of Equation \eqref{eq:sup_contr_and_cmmd} refers to the approximation error of the Taylor series used to obtain the equation. The term $ A - B/2 $ can be interpreted the difference of the similarity between all the features, and the similarity between features of the same class and domain. In a nutshell, the contrastive loss compute the contrast with respect to all pairs, and the CMMD the contrast with respect to pairs with the same class and domain. This difference is adjusted in Equation \eqref{eq:sup_contr_and_cmmd}. 

\subsection{Contrastive Learning and class-separability}

Extending on the work of \citet{li2021selfsupervised}, we can relate the contrastive losses (Supervised and NT-Xent) to an Inter-class MMD (IMMD) through the following lemma.

\begin{lemma}
\label{lemma:contr_loss_hsic_mmd_bound}

By assuming that the kernel $k$ is bounded, i.e., $| k(x,x') | < k^{max}$, $\forall x,x'$, and that the inner product on $\mathcal{Y} $ satisfies $ \langle y, y' \rangle_{\mathcal{Y}} = \Delta l \ \one_{\{ y = y' \} } + l_0 $, then the Contrastive losses bound the IMMD:

\begin{dmath}
    \label{eq:contr_loss_hsic_mmd_bound}
    - \frac{1}{\alpha} \text{IMMD}^2 + \gamma \text{HSIC}(X,X) + \mathcal{O} \left( \mathrm{Var} \left[ k \left( X, X' \right) \right] \right) \leqslant \mathcal{L}_{Contr},
\end{dmath}
with
\begin{dmath}
    \text{IMMD}^2 = \mathbb{E}_{C_1, C_2 \sim \pi^{\mathcal{Y}}_{0.5}} \left[ \| \mathbb{E}_{X \sim \pi^{\mathcal{X}|\mathcal{Y}}_{0.5, C_1}} [ \phi(X) ] \\ - \mathbb{E}_{X \sim \pi^{\mathcal{X}|\mathcal{Y}}_{0.5, C_2}} [ \phi(X) ] \|^2_{\mathcal{Z}}\right],
\end{dmath}
where $\text{HSIC}(X,X) $ is the Hilbert-Schmidt Independence Criterion \citep{gretton_hsic}, $\alpha$ is a proportionality constant which depends on problem parameters, and $\gamma \in \mathbb{R}$ is a constant satisfying $ \max \{ 2, 2 k^{max} \} = (1 + \sqrt{1 - 4 \gamma}) / (2 \gamma) $. For the Supervised Contrastive loss, $ \Delta l= K$ (the number of classes). For the NT-Xent loss, $ \Delta l = N$ (the number of instances).

\end{lemma}

The IMMD computes the difference between embeddings in the mixture domain with different class, and is thus a measure of class-separability. The HSIC is equal to the covariance in the feature space $\mathcal{Z}$, and it thus measures a non-linear covariance between the instances given the map $\phi$. We remark that the condition on $ \langle y, y' \rangle_{\mathcal{Y}} $ is satisfied when considering one-hot vectors and the Euclidean inner product on the label space. \citet{li2021selfsupervised} proved Lemma \ref{eq:contr_loss_hsic_mmd_bound} for the NT-Xent loss. In Appendix \ref{app:proof_contr_loss_hsic_mmd_bound} we extend the proof to the Supervised Contrastive loss.

To measure class-separability in the feature space we define another MMD-based quantity, the Different-class MMD (DCMMD), which is more general than the inter-class MMD of Equation \eqref{eq:contr_loss_hsic_mmd_bound}. The DCMMD measures the difference between the features of two different classes, in the same and different domains.

\begin{definition}[DCMMD] 
\label{def:dcmmd}

Given two labeled domains $\mathcal{D}_0$, $\mathcal{D}_1$, and a mixed domain $ \mathcal{D}_p $, the DCMMD is defined as:
\begin{dmath}
    \label{eq:dcmmd_def}
    \text{DCMMD}^2(\mathcal{D}_0, \mathcal{D}_1, \phi) = \mathbb{E}_{C_1, C_2 \sim \pi^{\mathcal{Y}}_{p_0}; C_1 \neq C_2; D_1,D_2 \sim Ber(p)} \\ \left[
    \left\| \mathbb{E}_{X \sim \pi^{\mathcal{X}|\mathcal{Y}}_{D_2,C_1}} [ \phi(X) ] - \mathbb{E}_{X \sim \pi^{\mathcal{X}|\mathcal{Y}}_{D_1,C_2}} [  \phi(X) ] \right\|_{\mathcal{Z}}^2
    \right],
\end{dmath}
where $Ber(p)$ is the Bernoulli distribution. To remain consistent with the CMMD definition, we usually consider equiprobable domains and set $p=1/2$.
    
\end{definition}

\section{Numerical experiments}

This section describes the datasets, models, and training settings used for the numerical experiments.

\subsection{Datasets}

Three types of mammography datasets are considered in this work: a clinical mammography image dataset (GEHC image dataset), a clinical mammography patch dataset (GEHC patch dataset), and a synthetic mammography patch dataset (synthetic patch dataset). The two clinical datasets contain GEHC images from anonymized patients, collected from a single institution in France following the EU General Data Protection Regulation. Additionally, two publicly-available datasets, CBIS-DDSM \cite{cbis_ddsm} and InBreast \cite{InBreastDataset} are also used in this work.

\paragraph{GEHC image dataset.} It contains 1300 cases, of which 197 are biopsy-proven cancers and 313 contain benign biopsied lesions. The remaining 790 are normal cases, which are studies in which no suspicious lesion was found in the breasts, and are confirmed by a one-year follow-up exam. The dataset is split in training (936 cases), validation (167 cases), and test (197) subsets in a stratified fashion, which takes into account the case pathology (benign or malignant), the lesions contained in the image (mass and calcification), and the description or sub-type of the lesions (e.g., spiculated mass, oval mass, granular calcification, etc.).

\paragraph{GEHC patch dataset.} Ten normal patches, and at least ten lesion $512 \times 512 $ pixel patches are extracted from each image that contains a lesion (mass or calcification), with two different strategies: ``fixed'' and ``random'' extractions. For every lesion, a ``fixed'' patch centered in the lesion centered is extracted. If the lesion is too large to be entirely contained in the patch, the space covered by the lesion is divided into a grid of $N \times M$ non-overlapping patches, which are incorporated to the patch dataset. This assures that every part of the lesion is represented in the dataset but may introduce an undesirable bias, as most patches coming from large lesions contain the lesion fragment in the corners. To reduce this bias, the patch dataset is enriched with ``random'' lesion patches, centered at random positions of the lesion. The extracted patches have an Intersection over Union (IoU) smaller than 0.5 between each other, to avoid generating patches that are too similar.

\paragraph{Synthetic patch dataset.} A synthetic patch dataset is created to enable controllable and efficient experiments while maintaining resemblance to real images. Mammography patches are generated by first sampling a Gaussian texture, and then inserting simulated mammography lesions. First, a white Gaussian random field $w \in \mathbb{R}^{N \times M}$ is sampled. Then, a low-pass filter with the following transfer function is applied to $w$:
\begin{equation}
 \label{eq:low_pass_filter}
    H(u, v) = \frac{1}{\sqrt{u^2+v^2}^\beta},
\end{equation}
where $u$ and $v$ are the coordinates of the image in the frequency space and $\beta$ is a non-negative real slope parameter, which can be associated to the breast density \cite{mainprize2012relationship}. The application of $H$ adds some spatial correlation to the pixels and creates the base texture. Two types of simple breast lesions are generated and inserted in the texture: masses and calcifications. Masses are simulated by randomly-centered Gaussian intensity profiles. Calcifications are modeled as high intensity pixels, clustered in random regions of the texture.

The synthetic patch dataset contains three types of patches: normal (only simulated breast texture), mass (breast texture containing an added synthetic mass), and calcification (breast texture with some synthetic calcifications), and it thus defines a three-class classification problem. A dataset of 1k 256 $\times$ 256 pixel synthetic patches, balanced in terms of classes, is generated. The parameters for generating this dataset are detailed in Appendix \ref{app:synthetic_params}. 

\begin{figure}[H]
    \centering
    \subfloat[Normal.]{
        \includegraphics[width=0.25\columnwidth]{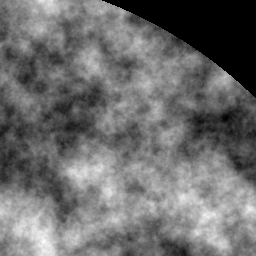}
    }
    \hspace{0.02\textwidth}
    \subfloat[Mass.]{
        \includegraphics[width=0.25\columnwidth]{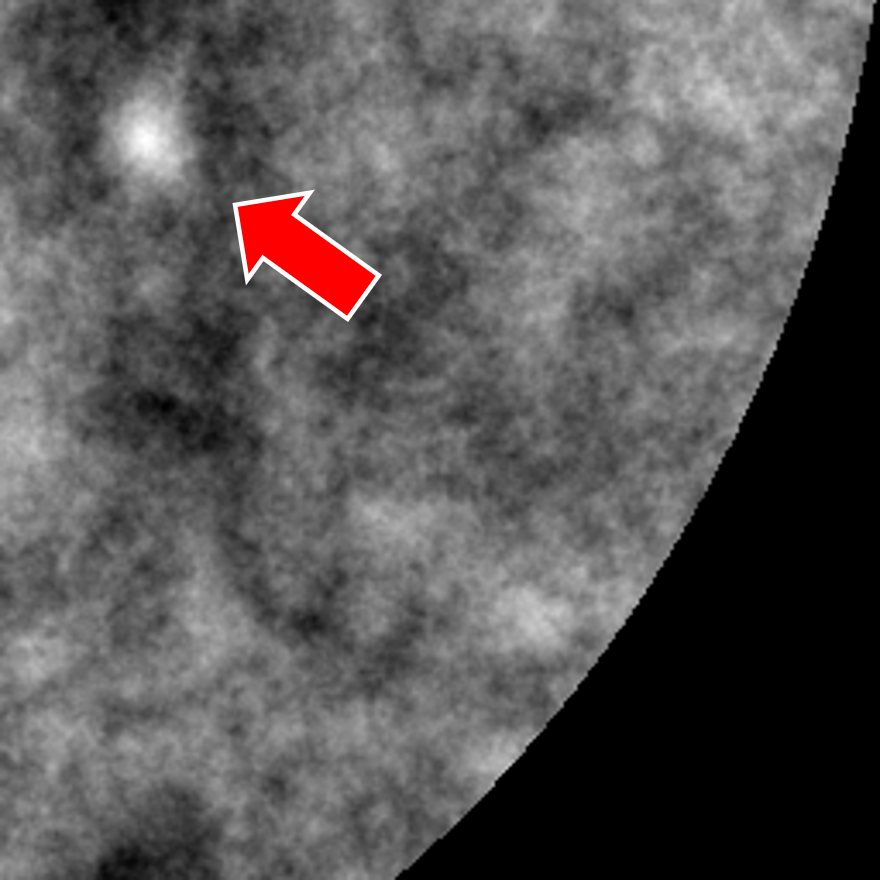}
    }
    \hspace{0.02\textwidth}
    \subfloat[Calcification.]{
        \includegraphics[width=0.25\columnwidth]{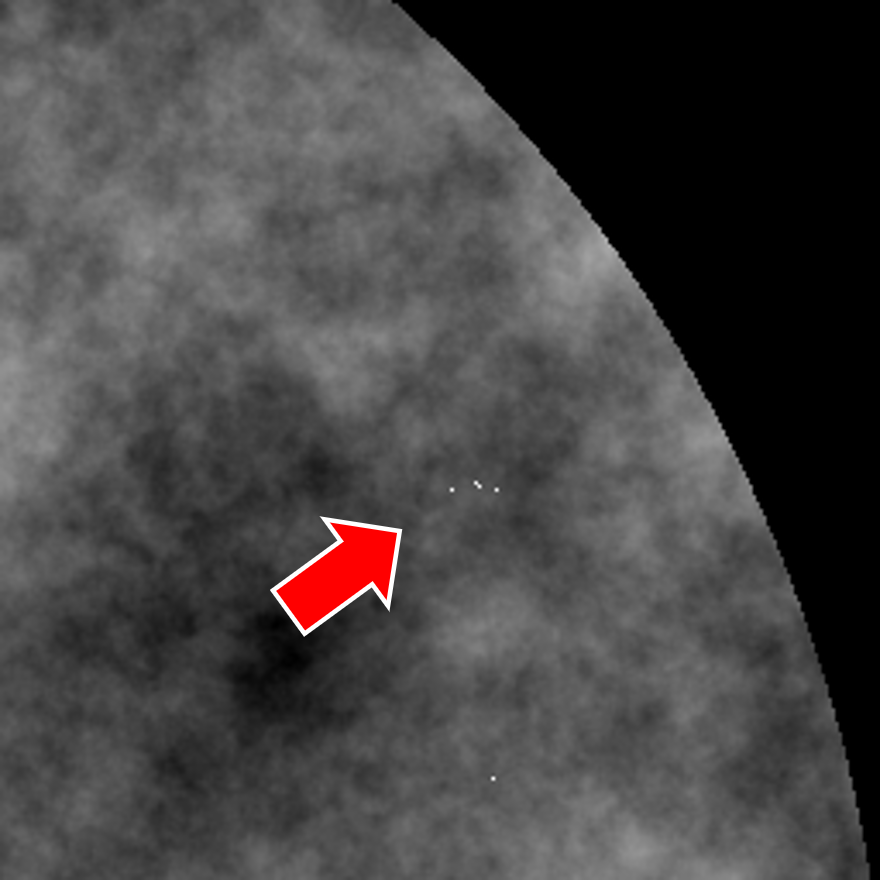}
    }
    \caption{\small Examples of synthetic patches (arrows signaling the lesions were inserted).}     
\label{fig:1-synthetic_patch_examples}
\end{figure}

\subsection{Image style heterogeneity}

\begin{figure}[ht]
  \centering

  \subfloat[\centering Without LUT]{\includegraphics[width=0.2\columnwidth]{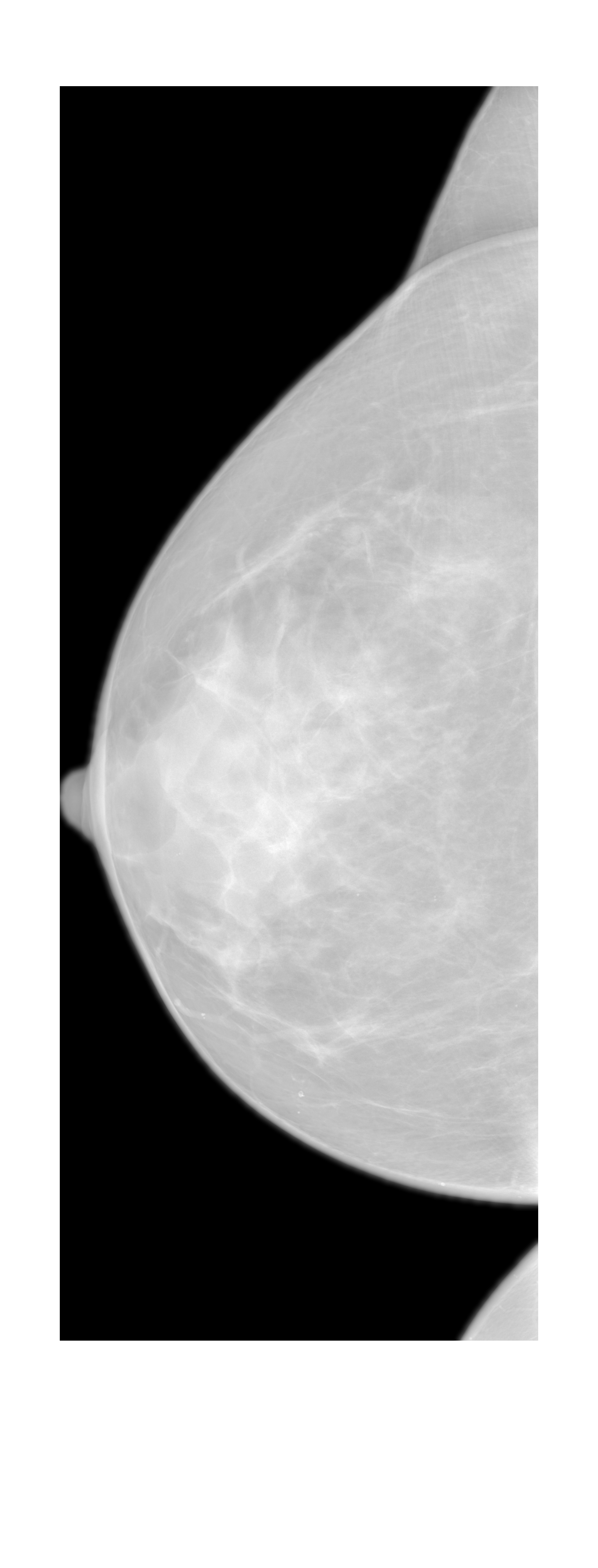}}
  \subfloat[\centering  With LUT]{\includegraphics[width=0.2\columnwidth]{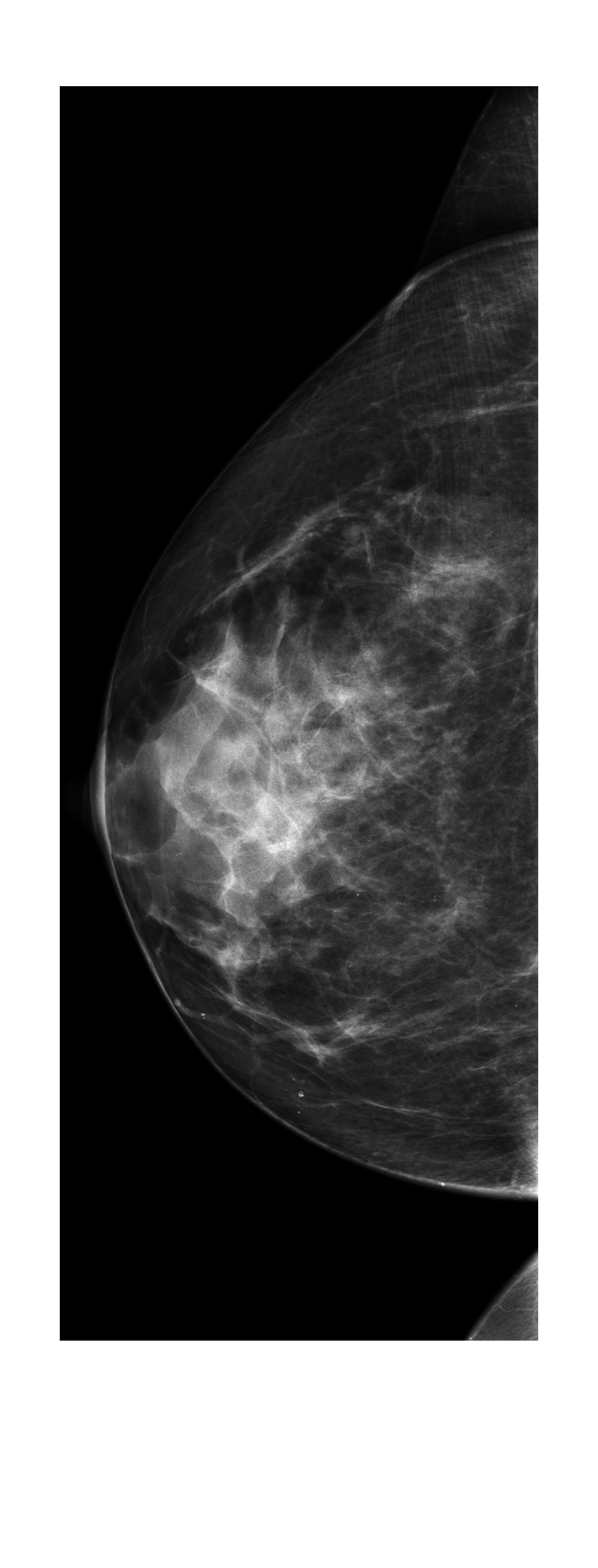}}
  \subfloat[\centering Pixel intensity histogram]{\includegraphics[width=0.5\columnwidth]{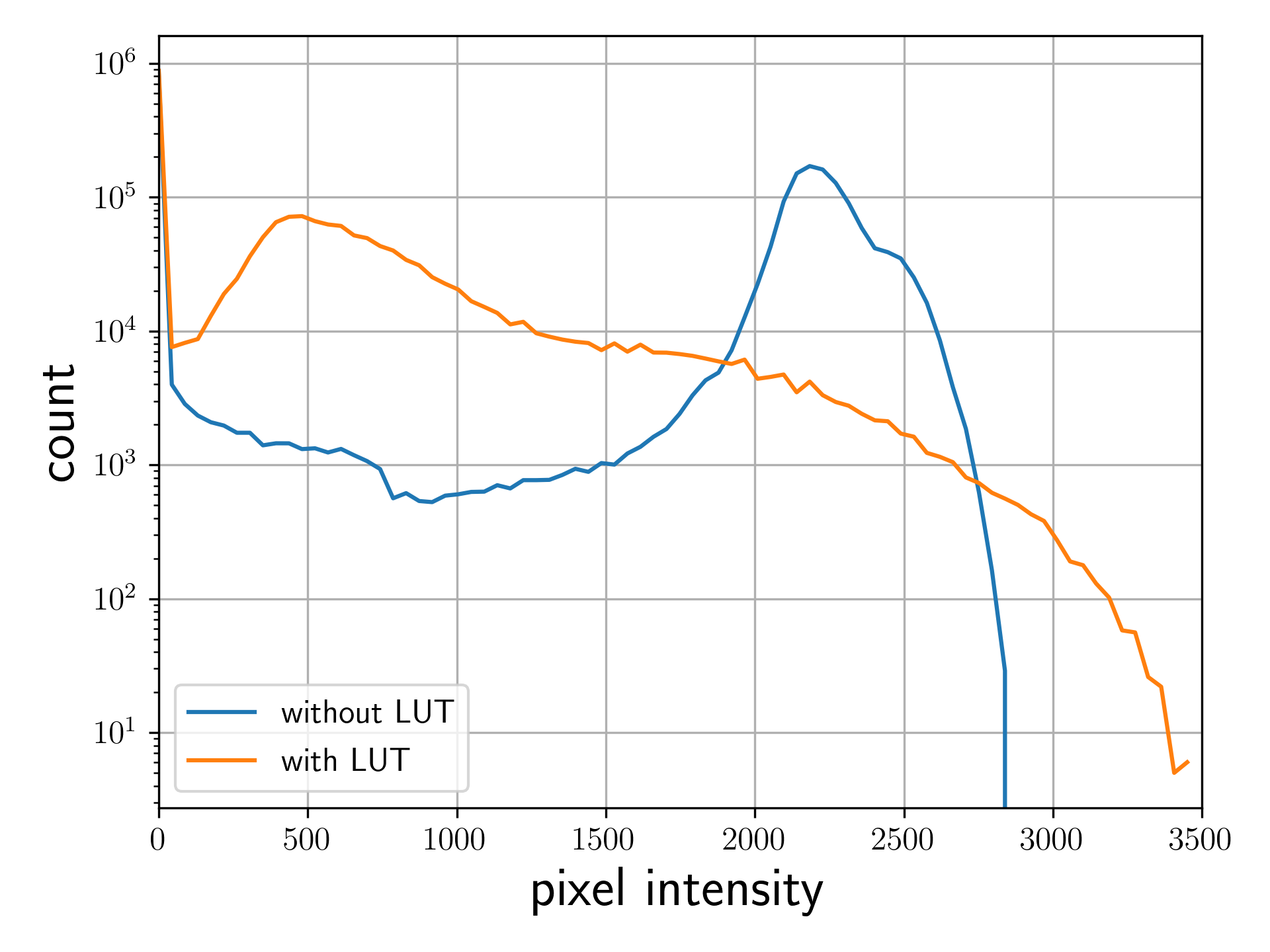}}

  \caption{\small Illustration of a single FFDM image: (a) without LUT application, (b) with LUT application, and (c) the pixel intensity histogram in logarithmic scale for both post-processings.}
  \label{fig:img_w_and_wo_lut_and_hst}
\end{figure}

In this work, we focus on studying the effect of training a DL model with data with different image styles. In particular, we use the sigmoid Look-Up Table (LUT) function, a contrast enhancement technique commonly used in mammography \cite{hernandez2024hybrid, breast_cancer_screening}, as proxy transformation to define the two data distributions or domains. However, the methodology developed in this work is applicable to any other image style or contrast transformation. Figure \ref{fig:img_w_and_wo_lut_and_hst} shows an example of the LUT application on a full mammography image.

Two distinct types of datasets are created from the base datasets for training and validation purposes, by splitting mammograms and patches at the case level:

\begin{itemize}
    \item An \textit{augmented dataset}, which includes two versions of each image: one with the LUT applied and the other without. This dataset can be seen as equivalent to applying a LUT-based data augmentation.
    \item \textit{Mixed datasets}, where the original dataset was divided into two groups. One group had images with the LUT applied, while the other did not.
\end{itemize}

For the clinical images and patches, four mixed datasets are constructed. For the synthetic patches, a single mixed dataset is constructed, due to the simplicity of the problem and the less variability observed in the results. Models are trained using each of these datasets, and evaluated in the same test set, which contained the two versions of each image (with and without LUT).

\subsection{Model architecture and training methodology}
\label{sec-da_centralized:model_arch_and_training}

\begin{figure}[h]
    \centering
    \includegraphics[width=\columnwidth]{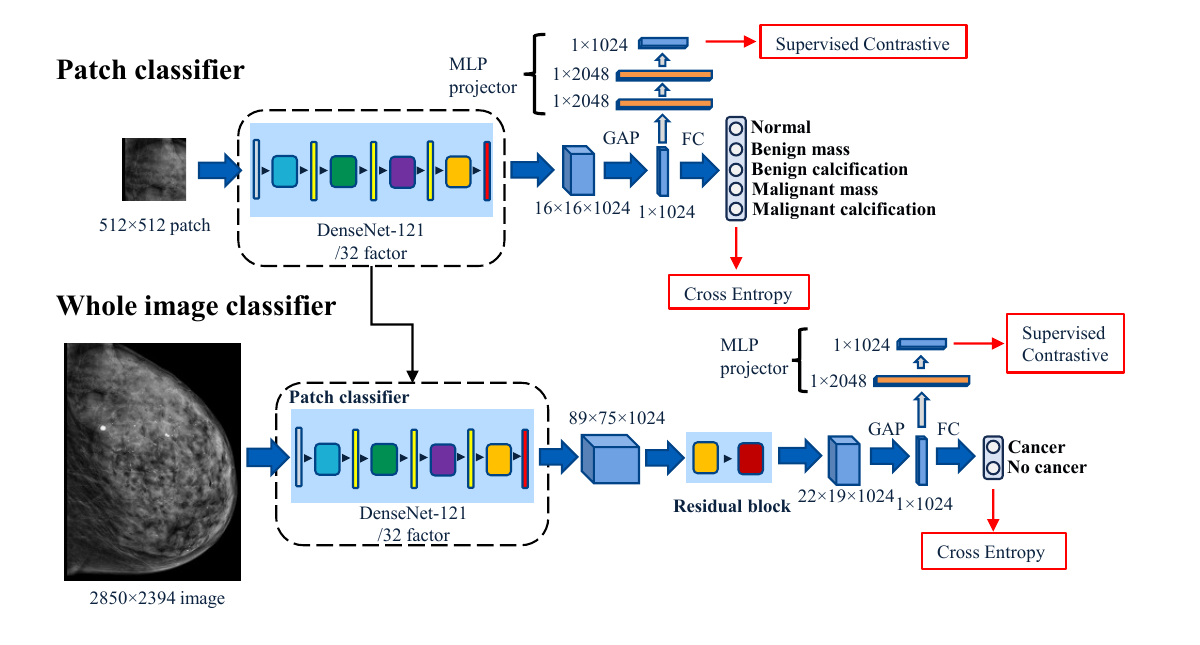}
    \caption{\small Patch-classifier and whole image classifier architectures for training with the Cross Entropy loss and with the Supervised Contrastive loss. GAP: Global Average Pooling, FC: Fully Connected layer.}
    \label{fig:1-model-arch-contrastive}
\end{figure}

A patch-based model is used for classifying mammography images, which consists in first training a path-classifier and then extending it to a whole image classifier by appending additional residual blocks and re-training on complete images (see Figure \ref{fig:1-model-arch-contrastive}). In this work, DenseNet-121 \citep{densenet} is used as backbone. For the clinical patch-classifier and whole image classifier, a Multi-layer Perceptron (MLP) projector is added for training with the Supervised Contrastive loss. The use of a projector is standard in CL, and enables to avoid the training task's overfitting bias \citep{SelfSupervisedLearningCookbook2023}, caused by the fact that the optimal features for the training (in this case, minimizing the Contrastive loss) task may not be optimal for the downstream task (in this case, classification). In the case of the patch-classifier, the projector features two hidden layers with 2048 units each, and an output layer of 1024 units (Figure \ref{fig:1-model-arch-contrastive} - top). For the whole image classifier, it consists of one hidden layer of 2048 units, and an output layer of 1024 units (Figure \ref{fig:1-model-arch-contrastive} - bottom). The projector is key to avoiding perfect invariance in the features used for classification, which lower the classification performance for clinical images, and it is used solely for the model training phase with Supervised Contrastive Learning (SCL), but not in the inference or evaluation stages. For synthetic patch classification the Supervised Contrastive loss is applied directly to the extracted features, as perfect invariance did not affect classification performance. This is likely due to the simplicity of the problem, which makes the projector unnecessary. To initialize the clinical patch-classifier, two methods are explored: one using the ImageNet dataset and the other using the CBIS-DDSM dataset. As CBIS-DDSM is a 2D mammography image dataset, it is more similar to the GEHC dataset than ImageNet, and is thus expected to provide a better initialization. For obtaining the whole image classifier, only the patch-classifier initialized on CBIS-DDSM is used. 

A model trained solely with the Cross Entropy (CE) loss, and thus without Domain Adaptation, was compared to a model trained using the Supervised Contrastive loss. As suggested by \citet{sup_contr_ler}, the model is first trained with the Supervised Contrastive loss to extract domain-invariant features. Then, the feature extraction layers are frozen, and only the final linear classification layer is trained with CE. This training strategy, also known as Linear Classification Protocol (LCP) is standard in Contrastive Learning \citep{MoCo, MICCAI2021_fed_2}. The resulting model is denoted as SupContr+LCP. As a third training strategy, the SupContr+LCP model is fully re-trained using the Cross Entropy loss (without freezing the feature extraction layers), resulting in a model denoted as SupContr+CE. These three training strategies (CE, SupContr+LCP, and SupContr+CE) are compared for patch, both synthetic and clinical, and whole image classification. The experimental settings and hyperparameters are detailed in Appendix \ref{app:exp_details}.

In addition, the generalization capability of the whole image classifier is assessed on InBreast, a publicly available dataset of mammography images. For this, the InBreast dataset is split into training (288 cases), validation (46 cases) and testing (75 cases) sets, with the same stratification strategy used for the GEHC dataset. To keep the learned representation fixed, the feature extractor is frozen during InBreast fine-tuning, and only the output linear layer is updated.

\begin{adjustwidth}{1.8cm}{1.8cm}
\begin{table*}[tb]
    {
    \scriptsize
    \begin{tabular}{p{2cm}|p{2.07cm}|p{2.07cm}|p{2.07cm}|p{2.07cm}|p{2.07cm}|p{2.07cm}}
    Model & CMMD $\downarrow$ & DCMMD $\uparrow$ & AUC $\uparrow$ & AUC (OvO) $\uparrow$ & AUC (OvR) $\uparrow$ & Accuracy $\uparrow$ \\
    \hline
    \multicolumn{6}{l}{Patch-classifier (synthetic - 3 classes)} & \\  
    CE & 0.348 $\pm$ 0.010 & 0.405 $\pm$ 0.005 & \centering - & 0.998 $\pm$ 0.002\textsuperscript{*} & 0.995 $\pm$ 0.005\textsuperscript{*} & \textbf{0.981 $\pm$ 0.018}\textsuperscript{*} \\
     SupContr+LCP & 0.239 $\pm$ 0.032 & 0.417 $\pm$ 0.010 & \centering - & \textbf{1.000 $\pm$ 0.000} & \textbf{0.999 $\pm$ 0.001} & \textbf{0.985 $\pm$ 0.015}\textsuperscript{*} \\
    SupContr+CE & \textbf{0.226 $\pm$ 0.029} & \textbf{0.438 $\pm$ 0.007} & \centering - & 0.998 $\pm$ 0.003\textsuperscript{*} & 0.996 $\pm$ 0.006\textsuperscript{*} & 0.969 $\pm$ 0.02 \\
    \hline
    \multicolumn{6}{l}{Patch-classifier (clinical - TL from ImageNet - 5 classes)} & \\  
    CE &  0.120 $\pm$ 0.041 & 0.094 $\pm$ 0.041 & \centering - & 0.728 $\pm$ 0.017 & 0.684 $\pm$ 0.020 &  0.391 $\pm$ 0.037 \\
     SupContr+LCP & \textbf{0.092 $\pm$ 0.016}\textsuperscript{*} & \textbf{0.185 $\pm$  0.020}\textsuperscript{*} & \centering - & \textbf{0.847 $\pm$ 0.011}\textsuperscript{*} & \textbf{0.793 $\pm$ 0.019}\textsuperscript{*} & \textbf{0.497 $\pm$ 0.044}\textsuperscript{*} \\
    SupContr+CE & \textbf{0.092 $\pm$ 0.016}\textsuperscript{*} & \textbf{0.183 $\pm$ 0.021}\textsuperscript{*} & \centering - & \textbf{0.846 $\pm$ 0.012}\textsuperscript{*} & \textbf{0.792 $\pm$ 0.020}\textsuperscript{*} & \textbf{0.508 $\pm$ 0.019}\textsuperscript{*} \\
    \hline
    \multicolumn{6}{l}{Patch-classifier (clinical - TL from CBIS-DDSM - 5 classes)} & \\  
    CE & 0.092 $\pm$0.020\textsuperscript{*}  & 0.205 $\pm$ 0.042 & \centering - & \textbf{0.915 $\pm$ 0.016}\textsuperscript{*} & \textbf{0.880 $\pm$ 0.014}\textsuperscript{*} & \textbf{0.627 $\pm$ 0.035}\textsuperscript{*} \\
     SupContr+LCP & 0.081 $\pm$ 0.019\textsuperscript{*} & 0.267 $\pm$ 0.011 & \centering - & 0.878 $\pm$ 0.014 & 0.845 $\pm$ 0.012 & 0.569 $\pm$ 0.029 \\
    SupContr+CE & \textbf{0.069 $\pm$ 0.009} & \textbf{0.278 $\pm$ 0.011} & \centering - & \textbf{0.918 $\pm$ 0.010}\textsuperscript{*} & \textbf{0.880 $\pm$ 0.008}\textsuperscript{*} & \textbf{0.628 $\pm$ 0.005}\textsuperscript{*} \\
    \hline
    \multicolumn{6}{l}{Whole image classifier (2 classes)} & \\  
    CE & 0.118 $\pm$ 0.029 & 0.087 $\pm$ 0.042 & 0.718 $\pm$ 0.043 & \centering - & \centering - & 0.609 $\pm$ 0.084 \\
     SupContr+LCP & \textbf{0.050 $\pm$ 0.007}\textsuperscript{*} & \textbf{0.174 $\pm$ 0.041}\textsuperscript{*} & \textbf{0.759 $\pm$ 0.033}\textsuperscript{*} & \centering - & \centering - & \textbf{0.674 $\pm$ 0.023}\textsuperscript{*} \\
    SupContr+CE & \textbf{0.049 $\pm$ 0.011}\textsuperscript{*} & \textbf{0.151 $\pm$ 0.061}\textsuperscript{*} & \textbf{0.776 $\pm$ 0.009}\textsuperscript{*} & \centering - & \centering - & \textbf{0.698 $\pm$ 0.046}\textsuperscript{*} \\    
    \end{tabular}
    }
    \caption{\small Results on the mixed datasets.}
    \label{tab:res_mixed_ds}
\end{table*}
\end{adjustwidth}

\begin{adjustwidth}{2cm}{2cm}
\begin{table*}[tb]
    {\scriptsize
    \begin{tabular}{p{2cm}|p{2.07cm}|p{2.07cm}|p{2.07cm}|p{2.07cm}|p{2.07cm}|p{2.07cm}}
    Model & CMMD $\downarrow$ & DCMMD $\uparrow$ & AUC $\uparrow$ & AUC (OvO) $\uparrow$ & AUC (OvR) $\uparrow$ & Accuracy $\uparrow$ \\
    
    \hline
    \multicolumn{6}{l}{Patch-classifier (clinical - TL from ImageNet - 5 classes)} & \\  
    
    CE & 0.152 $\pm$ 0.004 & 0.163 $\pm$ 0.018\textsuperscript{*} & \centering - & 0.871 $\pm$ 0.006 & 0.817 $\pm$ 0.010 & \textbf{0.547 $\pm$ 0.022}\textsuperscript{*}  \\
     SupContr+LCP & \textbf{0.027 $\pm$ 0.005} & \textbf{0.229 $\pm$ 0.005}  & \centering - & 0.878 $\pm$ 0.005 & \textbf{0.846 $\pm$ 0.010} & 0.538 $\pm$ 0.020 \\
    SupContr+CE & 0.037 $\pm$ 0.006 & 0.171\textsuperscript{*} $\pm$ 0.004 & \centering - & \textbf{0.887 $\pm$ 0.005} & 0.805 $\pm$ 0.011 & \textbf{0.542 $\pm$ 0.021}\textsuperscript{*}  \\
    
    \hline
    \multicolumn{6}{l}{Patch-classifier (clinical - TL from CBIS-DDSM - 5 classes)} & \\  

    CE & 0.091 $\pm$ 0.004 & 0.226 $\pm$ 0.002 & \centering - & \textbf{0.927 $\pm$ 0.004} & \textbf{0.896 $\pm$ 0.007} & \textbf{0.656 $\pm$ 0.021} \\
     SupContr+LCP & \textbf{0.034 $\pm$ 0.004}\textsuperscript{*} & 0.243 $\pm$ 0.003 & \centering - & 0.880 $\pm$ 0.005 & 0.842 $\pm$ 0.009 & 0.551 $\pm$ 0.019 \\
    SupContr+CE & \textbf{0.032 $\pm$ 0.006}\textsuperscript{*} & \textbf{0.283 $\pm$ 0.003} & \centering - & 0.919 $\pm$ 0.004 & 0.881 $\pm$ 0.008 & 0.599 $\pm$ 0.020 \\
    
    \hline
    \multicolumn{6}{l}{Whole image classifier (GEHC dataset - 2 classes)} & \\ 
    
    CE & 0.108 $\pm$ 0.009 & 0.099 $\pm$ 0.018 & 0.745 $\pm$ 0.050 & \centering - & \centering - & 0.625 $\pm$ 0.061 \\
     SupContr+LCP & \textbf{0.040 $\pm$ 0.016}\textsuperscript{*} & 0.127 $\pm$ 0.022 & 0.763 $\pm$ 0.058 & \centering - & \centering - &  \textbf{0.671 $\pm$ 0.043}\textsuperscript{*} \\
    SupContr+CE & \textbf{0.066 $\pm$ 0.029}\textsuperscript{*} & \textbf{0.213 $\pm$ 0.030} & \textbf{0.816 $\pm$ 0.042} & \centering - & \centering - & \textbf{0.728 $\pm$ 0.073}\textsuperscript{*} \\   
 
    \end{tabular}
    }
    \caption{\small Results on the augmented datasets.}
    \label{tab:res_aug_ds}
\end{table*}
\end{adjustwidth}

For synthetic patches, the CE, SupContr+LCP, and SupContr+CE models are trained on the mixed synthetic dataset. For clinical patches and full images, the models are trained on the four mixed datasets, and on the augmented dataset. The results of the four trainings on the mixed datasets are aggregated to calculate a mean performance and 95\% Confidence Intervals (CI), as well as p-values¸of a one-sided Welch's t-test. For the synthetic patch-classifier and the trainings on the clinical augmented datasets, Bootstrapping was used for obtaining the 95\% CI and p-values.

\section{Results}

The numerical results are organized into three sections: first, an illustration of Lemma \ref{lemma:sup_contr_and_cmmd}, followed by a quantitative analysis, and finally, a qualitative analysis.

\subsection{Illustration of Lemma \ref{lemma:sup_contr_and_cmmd}}

\begin{figure}[ht]
    \centering
    \includegraphics[width=0.6\columnwidth]{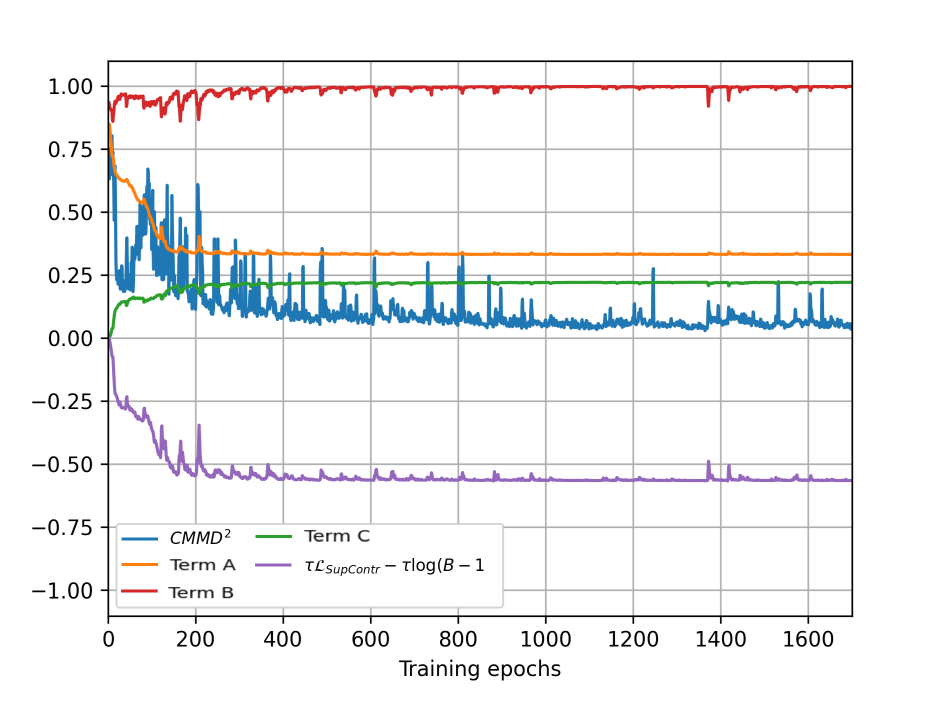}
    \caption{\small Evolution of the terms of Equation \eqref{eq:sup_contr_and_cmmd} during training.}
    \label{fig:evol-terms}
\end{figure}

Figure \ref{fig:evol-terms} shows the evolution of the individual terms in Equation \eqref{eq:sup_contr_and_cmmd} during training with synthetic patches. We observe that the CMMD and the similarity between all pairs, given by term A, decrease while minimizing the Supervised Contrastive loss, following Equation \eqref{eq:sup_contr_and_cmmd}. The similarity between pairs of features with the same class and domain increases, as predicted by Equation \eqref{eq:sup_contr_and_cmmd}. 

\begin{figure}[ht]
    \centering
    \includegraphics[width=0.6\columnwidth]{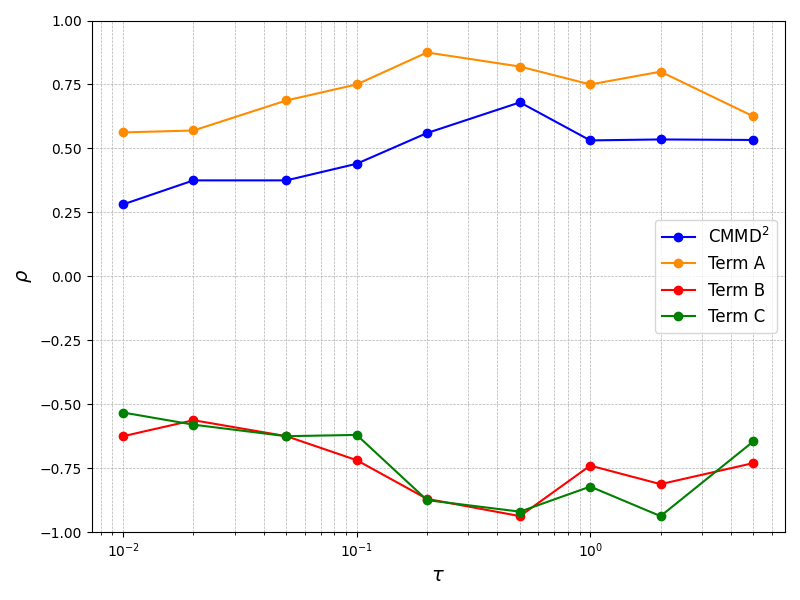}
    \caption{\small Evolution of the Pearson coefficients with the temperature.}
    \label{fig:3-deriv_correlation_vs_tmp}
\end{figure}

To quantify the trends observed, we analyze the correlation between the derivatives of each of the terms, and the derivative of the Supervised Contrastive loss for different temperature values $\tau$. This enables to numerically assess if the quantities are moving in the same, or opposite direction to the Supervised Contrastive loss. Figure \ref{fig:3-deriv_correlation_vs_tmp} shows the evolution of the Pearson correlation coefficient of the derivatives $\rho$ with the temperature $\tau$ for different temperature values, spanning from $\tau = 0.01 $ to $\tau = 5.0$. While the Pearson correlation coefficient for the CMMD and term A is positive, which confirms that the two quantities move in the same direction, it is negative for terms B and C. We observe that the magnitude of the Pearson correlation coefficient increases with increasing temperature, and reaches a \textit{plateau} between $\tau = 0.2$ and $\tau = 0.5$. This is explained by the fact that the Taylor approximation used for proving Lemma \ref{lemma:sup_contr_and_cmmd} is valid under relatively large temperatures. However, it is important to note that a certain level of correlation is observed, even at lower temperatures.

\subsection{Quantitative analysis}

Table \ref{tab:res_mixed_ds} shows the results for the mixed datasets. Domain Adaptation is measured in terms of the CMMD and class-separability is measured by the DCMMD. Classification performance is evaluated with the accuracy and AUC for the binary whole image classifier, and with the accuracy, One vs. One (OvO) AUC, and One vs. Rest (OvR) AUC for the patch classifiers. We observe that in all the classification problems, the models trained with the Supervised Contrastive loss (SupContr+LCP, SupContr+CE) achieve higher Domain Adaptation and class-separability than the CE models. This translates into a higher downstream classification performance for the whole image classifier and patch-classifier with synthetic patches, and clinical patches with TL from ImageNet. On the contrary, when weights are initialized from CBIS-DDSM (scanned mammography films), SupContr+CE matches CE performance but fails to outperform it. In this case, the pre-training dataset is closer to the GEHC dataset than ImageNet, leading to improved domain adaptation, class separability, and classification performance. This reduces the negative impacts of fine-tuning with images from different domains.

Table \ref{tab:res_aug_ds} presents the results for the augmented datasets under consideration, demonstrating that the conclusions drawn from the mixed datasets remain valid. In addition, by comparing the two tables for the whole image classifier, we can see that the contrastive-based models trained on the mixed datasets outperform the CE model trained on the augmented dataset, despite the latter having been trained on twice the amount of data.

Finally, Table \ref{res:inbreast} shows the results on the publicly available InBreast dataset. The SupContr+CE model exhibits superior generalization by achieving a 13\% AUC increase with respect to the CE model. We argue that this is a measure of generalization capabilities of the representations, as images from InBreast are only used for fine-tuning the linear classification layer while freezing the feature extraction.

\begin{figure}

    \centering 

    \subfloat{ 
        \includegraphics[width=\columnwidth]{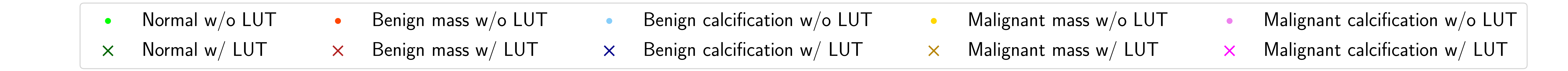} 
        \captionsetup[subfigure]{labelformat=empty}
        \addtocounter{subfigure}{-1}
        \vspace{-10cm}
    }

    \vspace{-0.5cm}
    \setcounter{subfigure}{0}

    \subfloat[ \small Transfer Learning from ImageNet. ]{%
        \label{fig:features_patch_clsf_imagenet}%
        \captionsetup[subfigure]{labelformat=empty}
        \subfloat[CE]{\includegraphics[width=0.32\columnwidth]{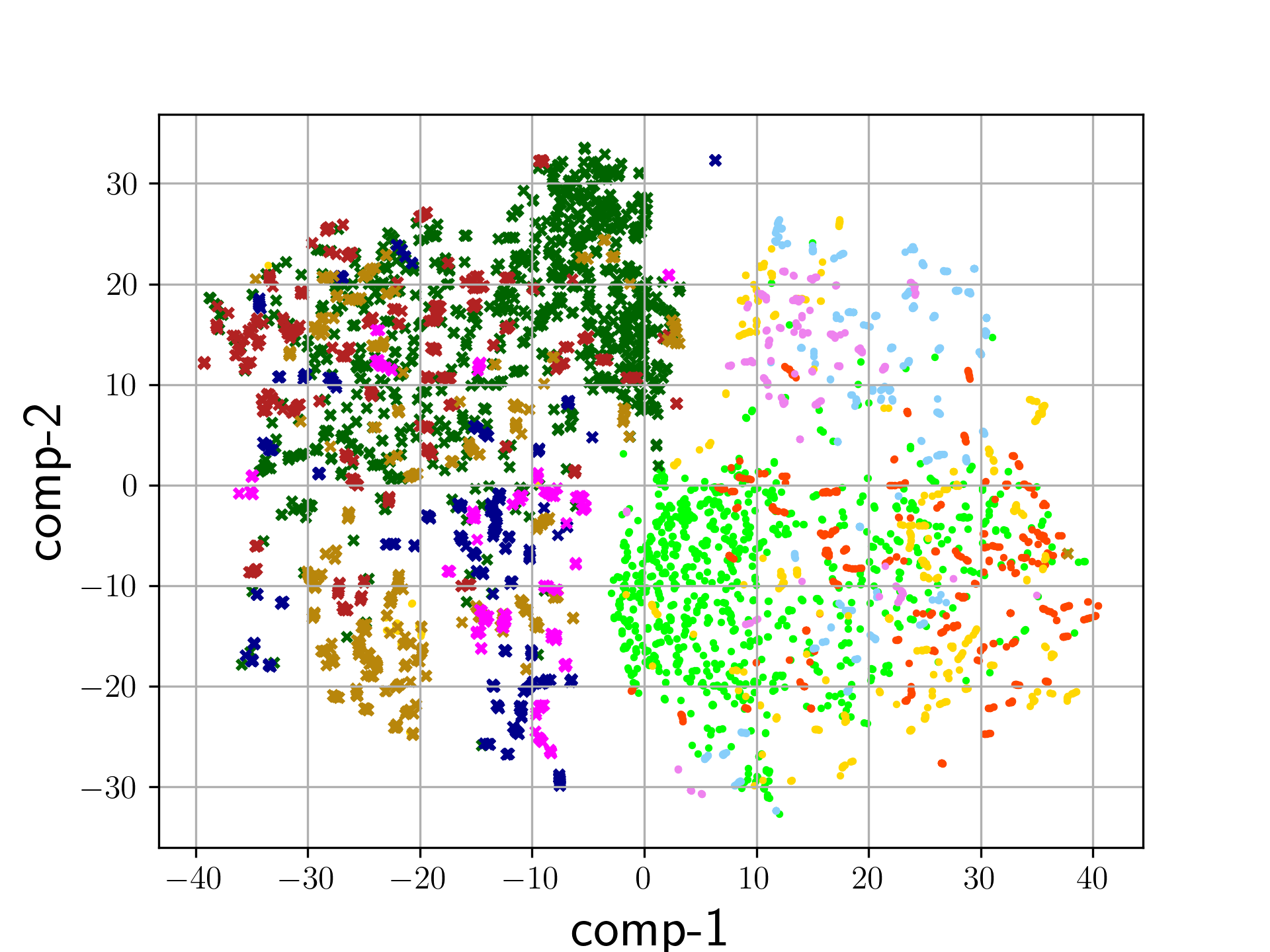}}
        \addtocounter{subfigure}{-1}
        \subfloat[SupContr+LCP]{\includegraphics[width=0.32\columnwidth]{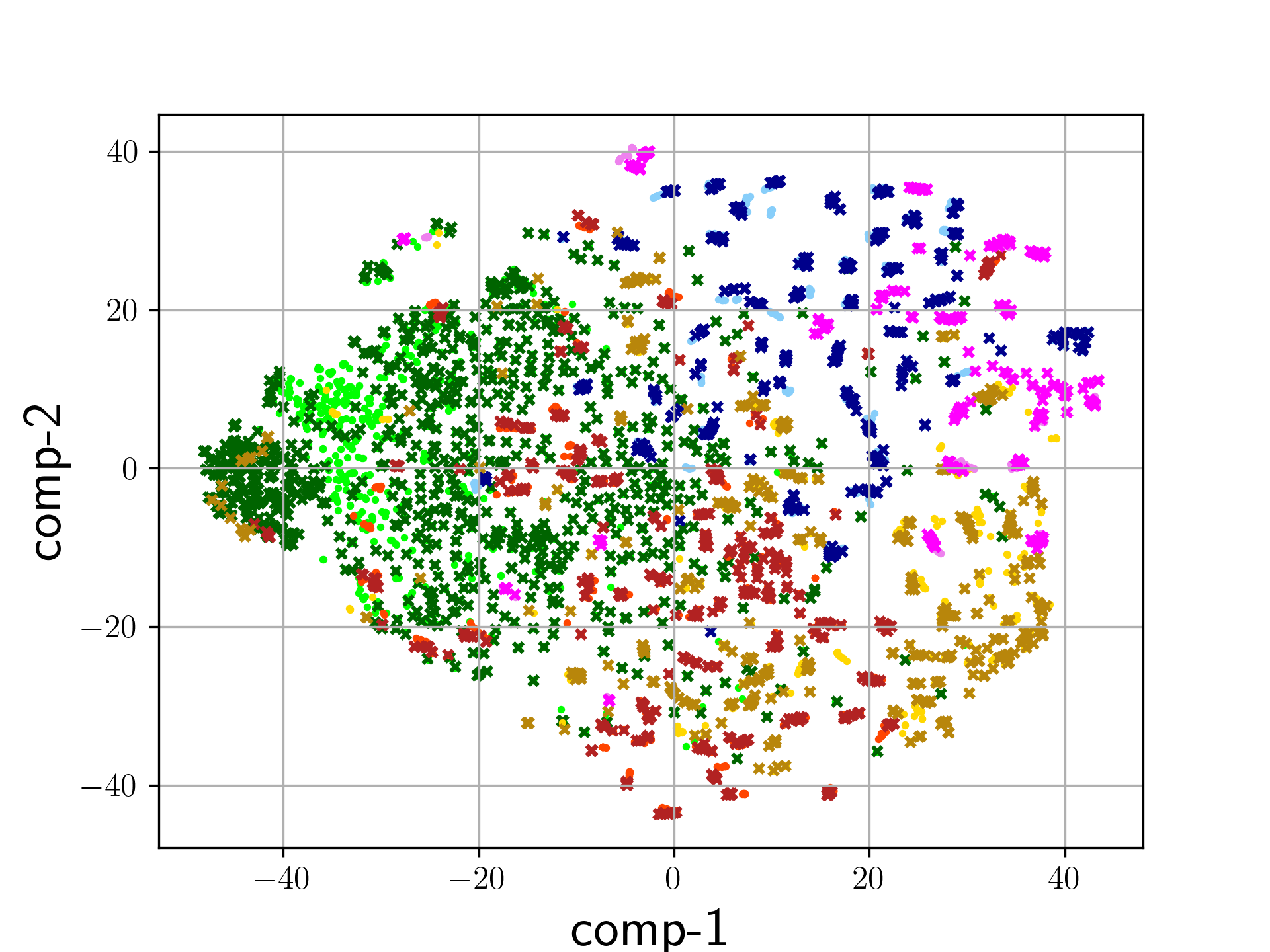}}
        \addtocounter{subfigure}{-1}
        \subfloat[SupContr+CE]{\includegraphics[width=0.32\columnwidth]{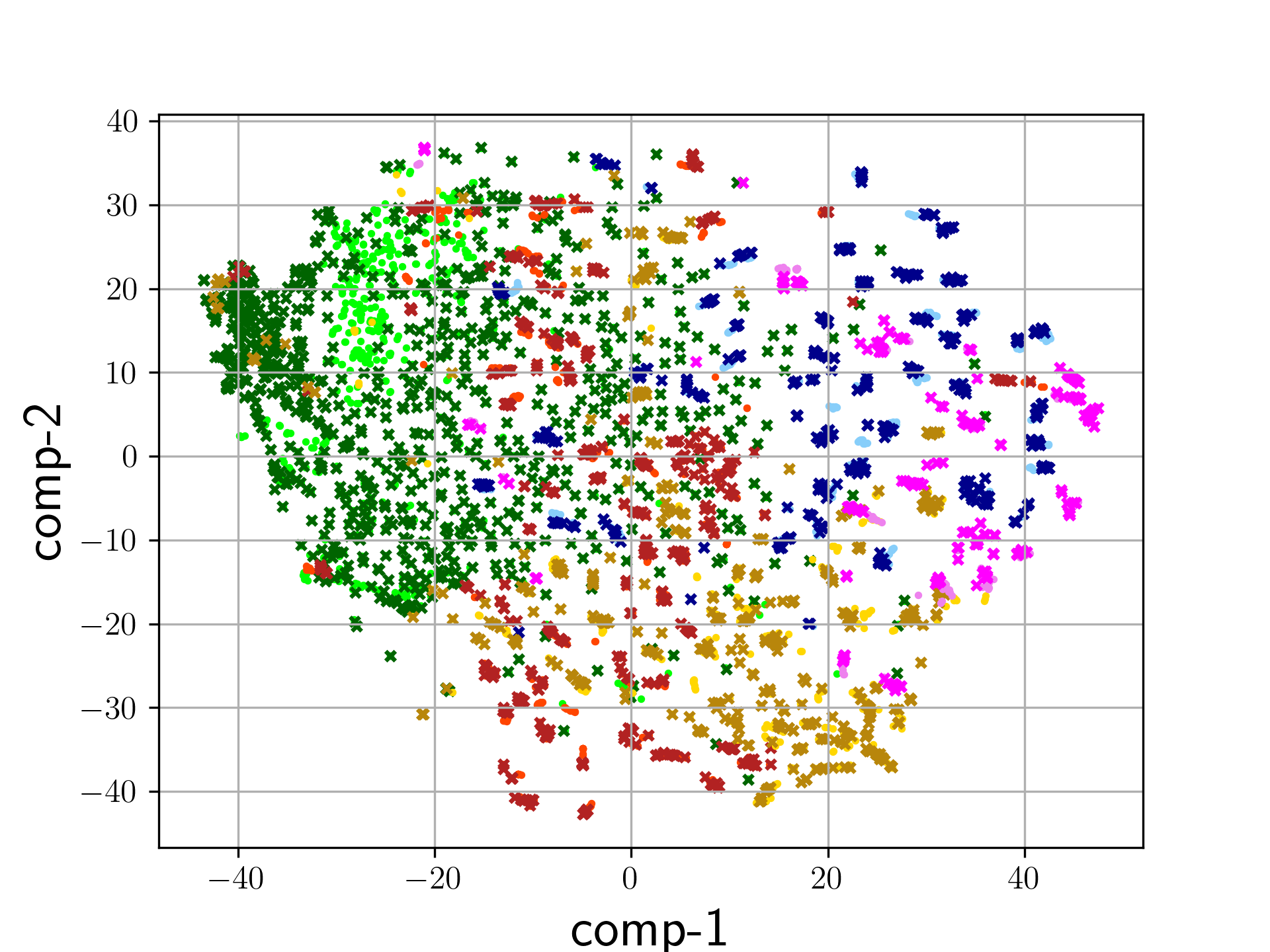}}
        \addtocounter{subfigure}{-1}
    }

    \vspace{-0.5cm}

    \subfloat[\small Transfer Learning from  CBIS-DDSM.]{%
        \label{fig:features_patch_clsf_ddsm}%
        \captionsetup[subfigure]{labelformat=empty}
        \subfloat[CE]{\includegraphics[width=0.32\columnwidth]{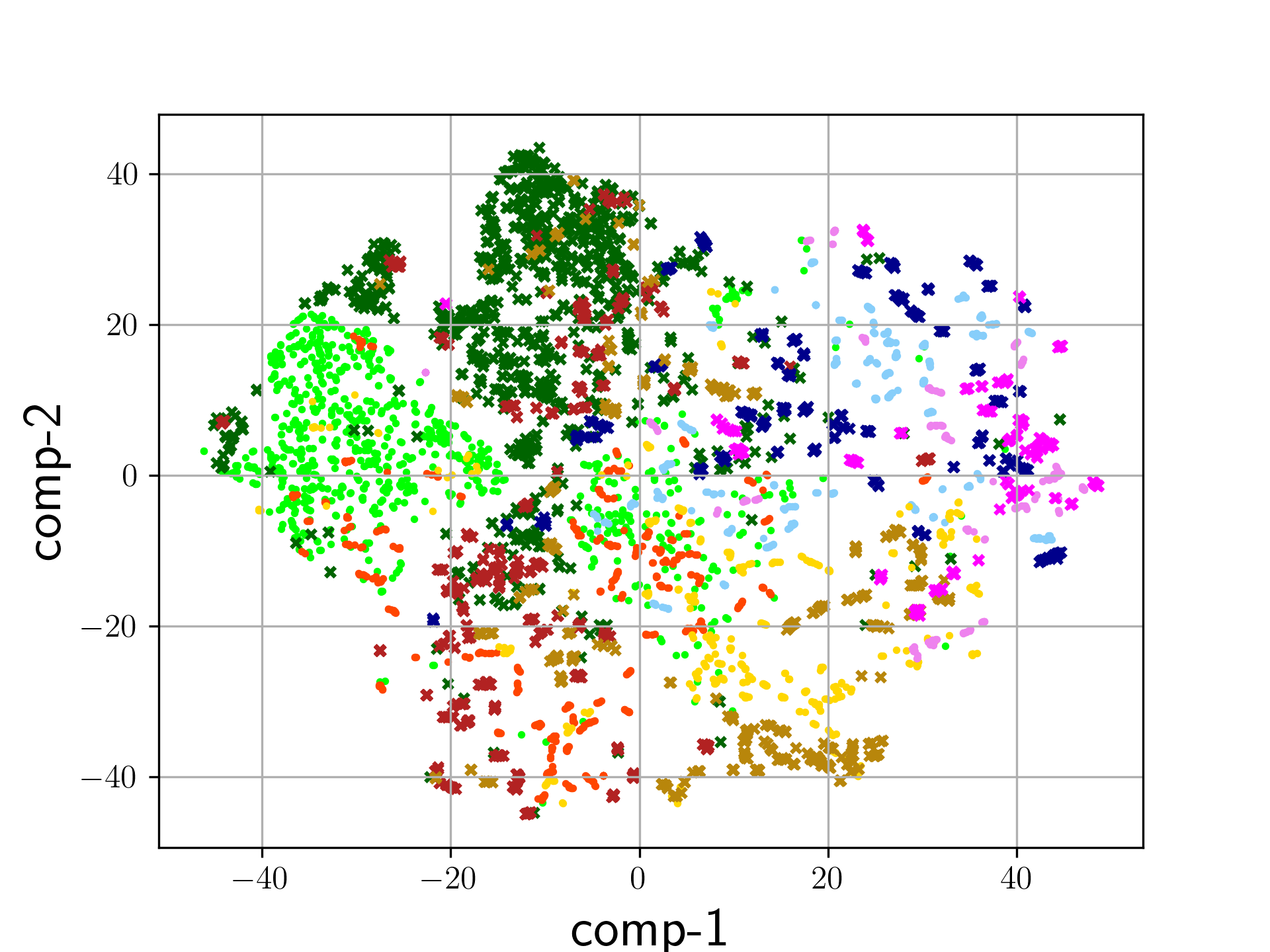}}
        \addtocounter{subfigure}{-1}
        \subfloat[SupContr+LCP]{\includegraphics[width=0.32\columnwidth]{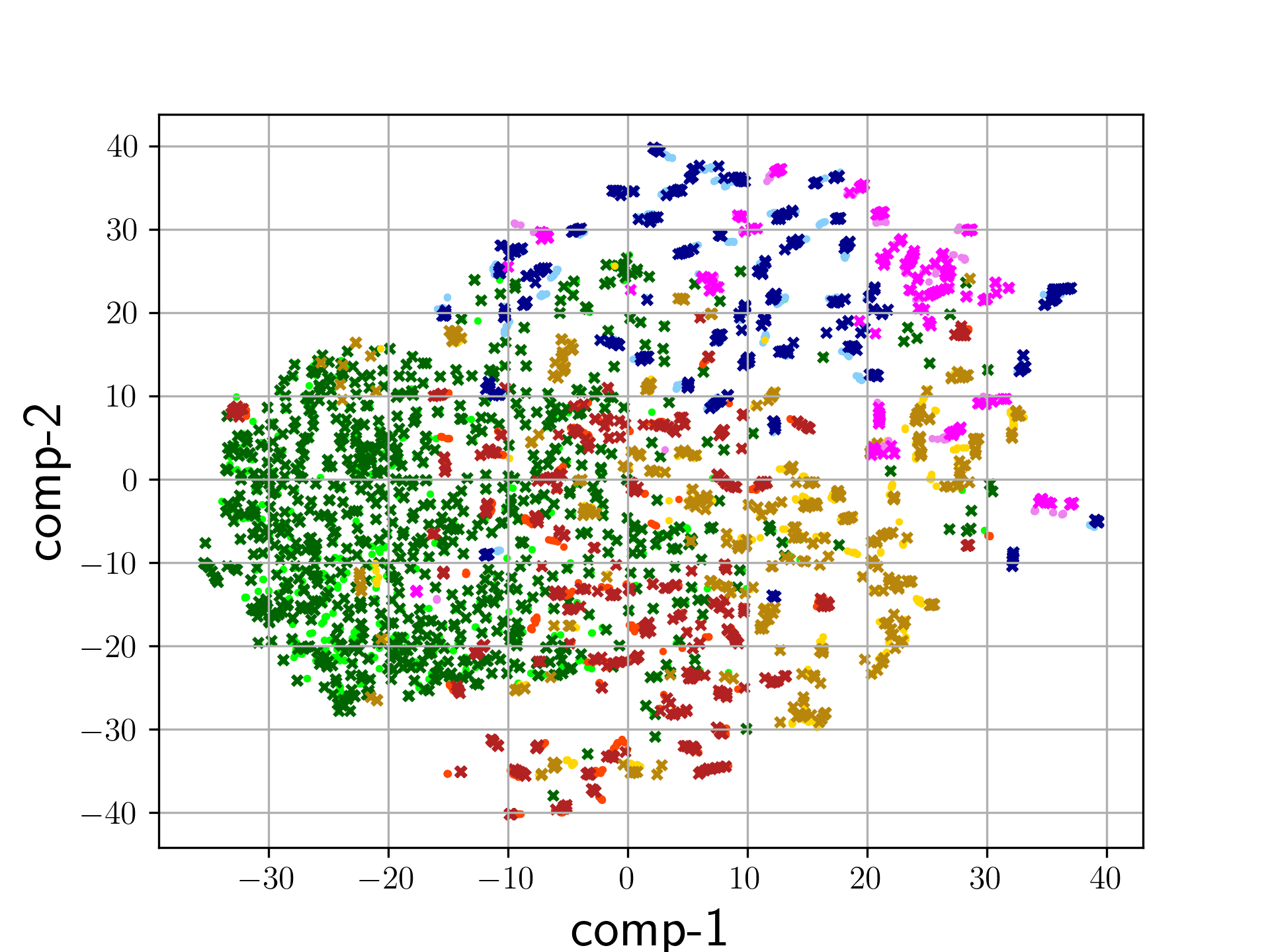}}
        \addtocounter{subfigure}{-1}
        \subfloat[SupContr+CE]{\includegraphics[width=0.32\columnwidth]{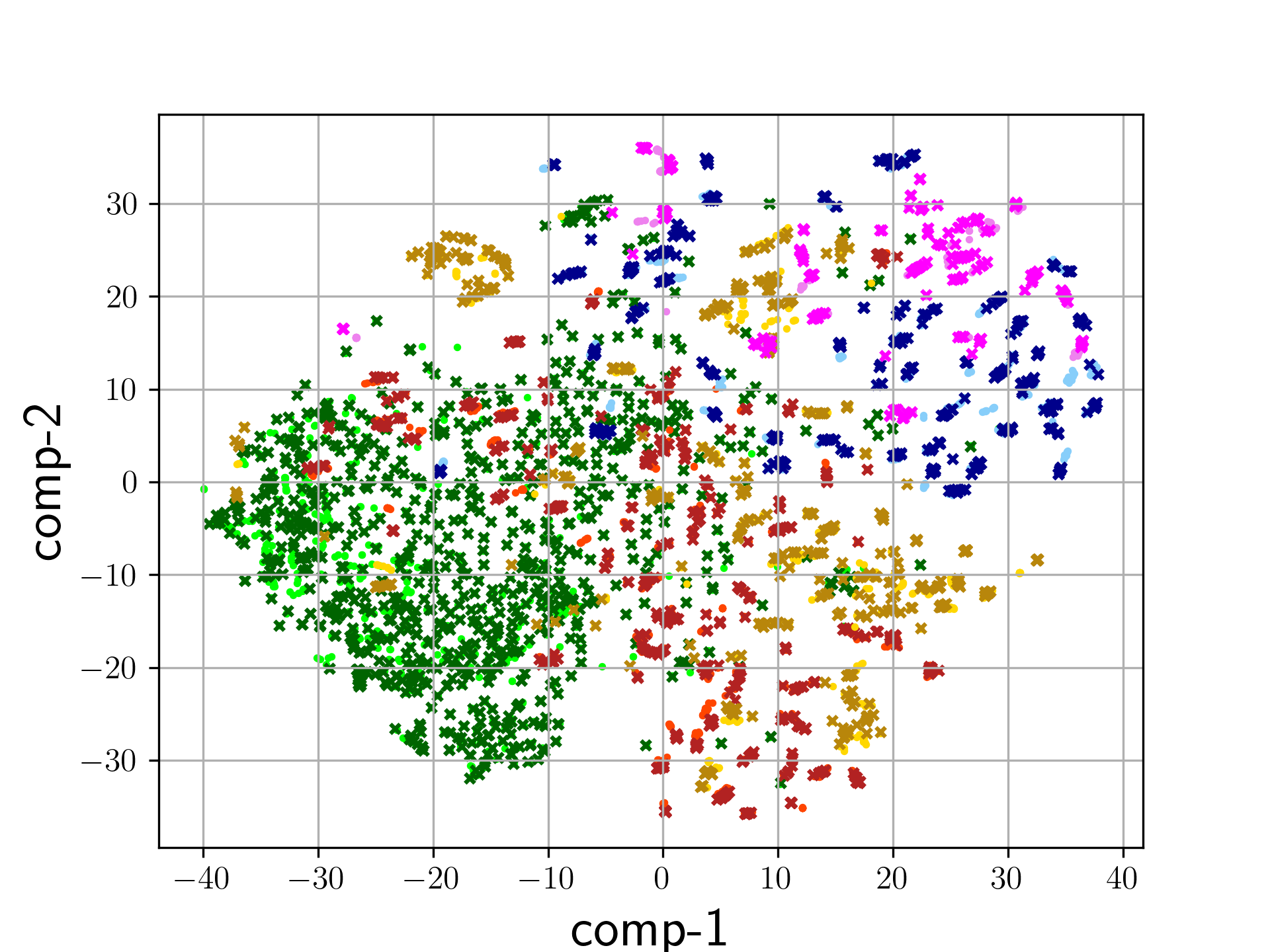}}  
        \addtocounter{subfigure}{-1}
    }
    
\caption{\small t-SNE plots of the features from the patch-classifier, indicating class and domain.}
\label{fig:tsne_features_patch_classifier_clinical}
\end{figure}

\begin{table}[htb]
    \centering
    \scriptsize
    \begin{tabular}{c|c|c}
         Model & AUC & Accuracy \\
         \hline
         CE & 0.733 $\pm$ 0.096\textsuperscript{*} & 0.571 $\pm$ 0.067 \\
         SupContr+LCP & 0.746 $\pm$ 0.083\textsuperscript{*} & 0.647 $\pm$ 0.061 \\
         SupContr+CE & \textbf{0.831 $\pm$ 0.071} & \textbf{0.703 $\pm$ 0.060} \\
    \end{tabular}
\caption{Results on InBreast.}
\label{res:inbreast}
\end{table}

\subsection{Qualitative analysis}

We now perform a qualitative evaluation of the extracted feature space. Figure \ref{fig:tsne_features_patch_classifier_clinical} shows the t-SNE plot of the extracted features for clinical patch-classifier, trained with the three losses (CE, SupContr+LCP, SupContr+CE), with weights initialized from ImageNet (Figure \ref{fig:features_patch_clsf_imagenet}) and CBIS-DDSM (Figure \ref{fig:features_patch_clsf_ddsm}). When Transfer Learning from ImageNet is used, the CE model features are more separated by domain than by class, while the SupContr+LCP and SupContr+CE models are domain-invariant (Figure \ref{fig:features_patch_clsf_imagenet}). When Transfer Learning from CBIS-DDSM is used (Figure \ref{fig:features_patch_clsf_ddsm}) it can be seen that the CE model has already some degree of domain invariance, especially for non-normal classes. We hypothesize that, in this case, the similarity of the CBIS-DDSM dataset to the GEHC images is leveraged by the DL-model, virtually increasing the training dataset and increasing the robustness of the learned features. As will be seen later, this decreases the impact of Domain Adaptation on classification performance. The SupContr+LCP and SupContr+CE models attain domain invariance for all the classes, including Normal patches.

\begin{figure}
    \centering

    \subfloat{
        \includegraphics[ width=1.05\columnwidth]{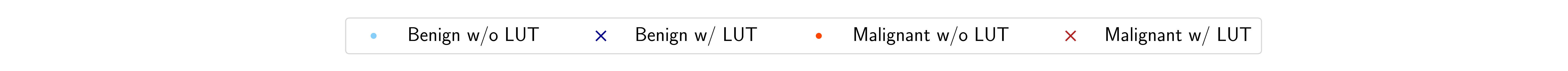} 
        \captionsetup[subfigure]{labelformat=empty}
        }
    \addtocounter{subfigure}{-1}
    \vspace{-1cm}

    \subfloat{
        \captionsetup[subfigure]{labelformat=empty}
        \subfloat[CE]{\includegraphics[width=0.32\columnwidth]{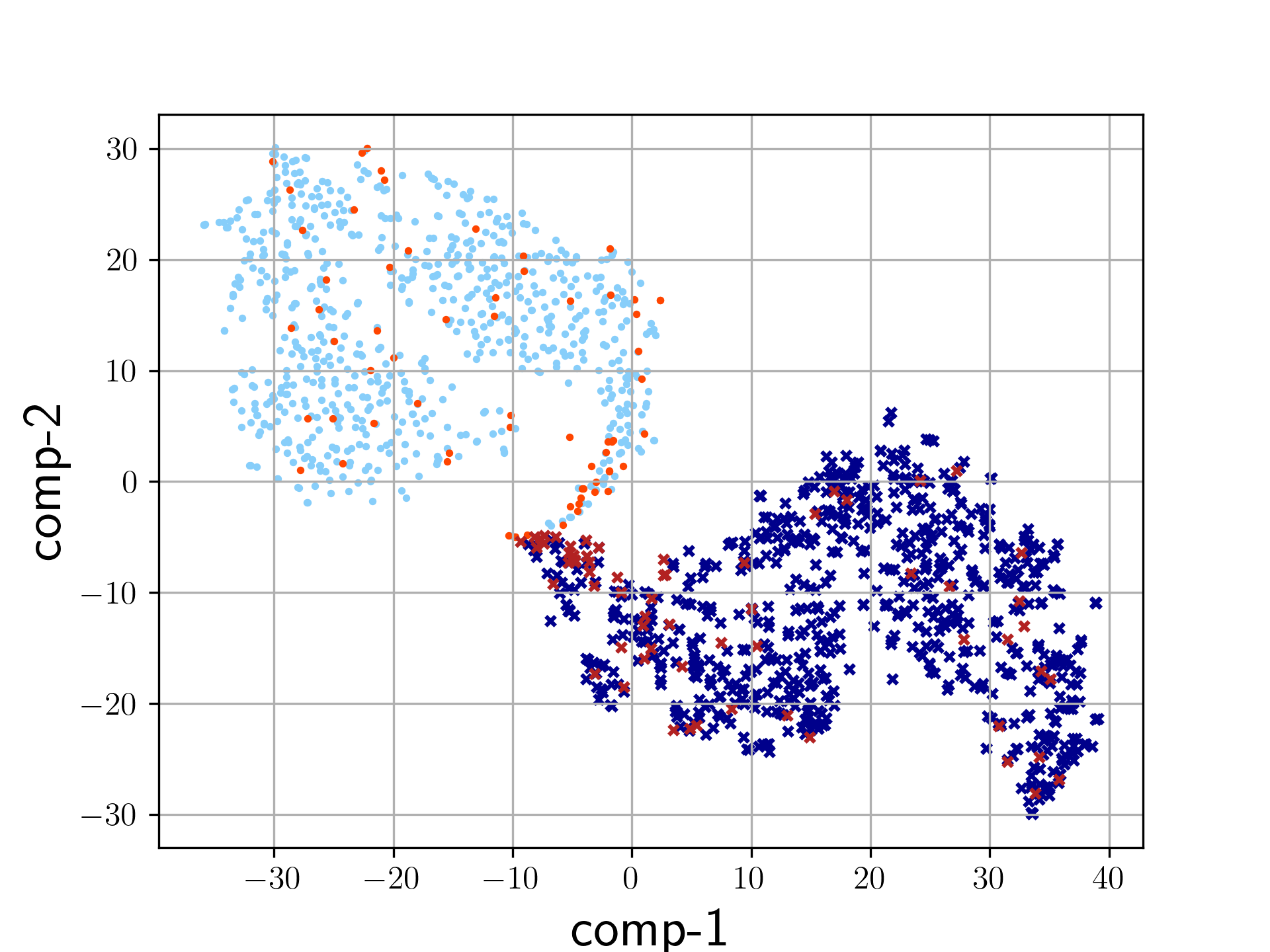}}
        \captionsetup[subfigure]{labelformat=empty}
        \addtocounter{subfigure}{-1}
        \subfloat[SupContr+LCP]{\includegraphics[width=0.32\columnwidth]{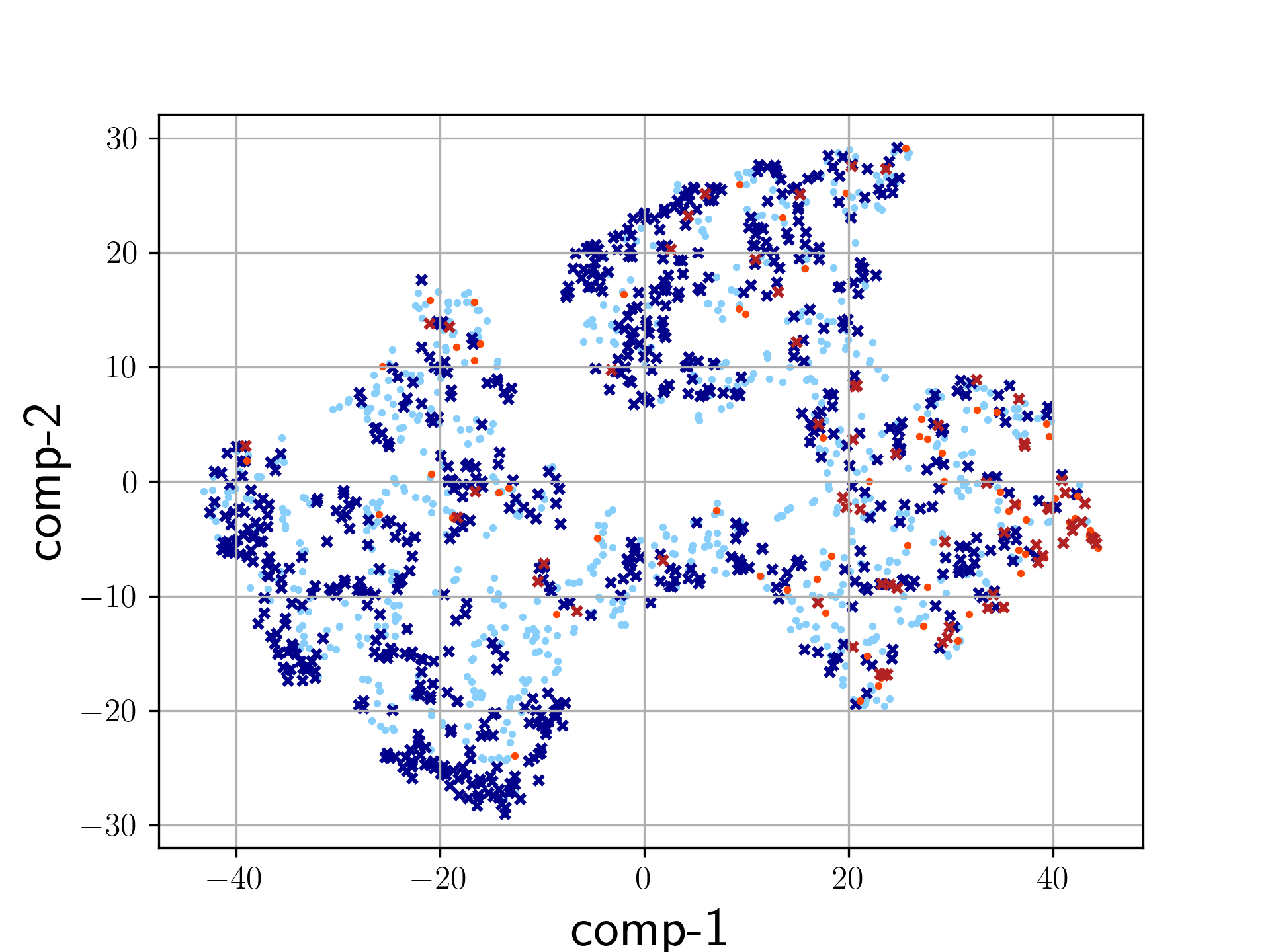}}    \captionsetup[subfigure]{labelformat=empty}
        \addtocounter{subfigure}{-1}
        \subfloat[SupContr+CE]{\includegraphics[width=0.32\columnwidth]{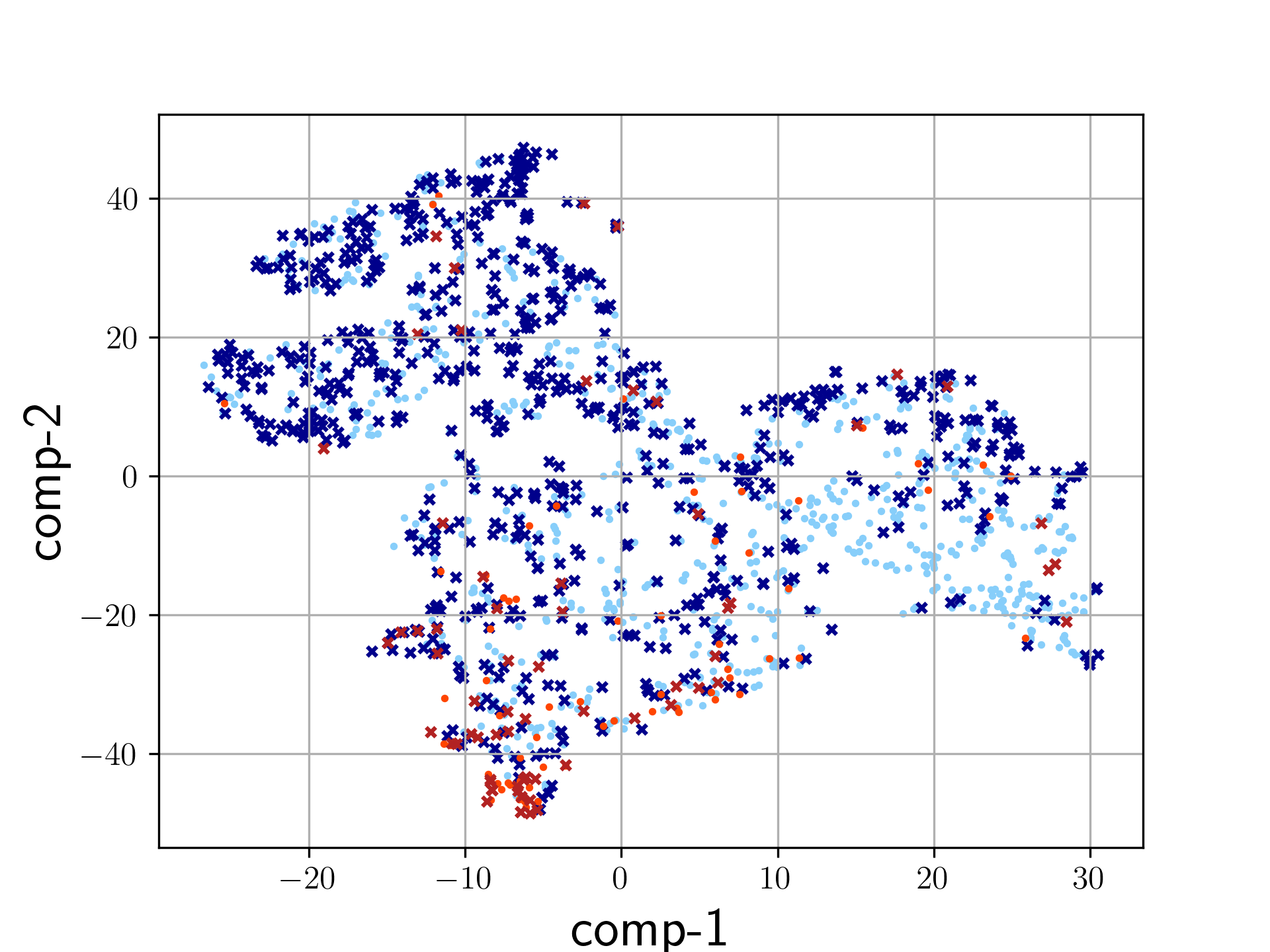}}    \captionsetup[subfigure]{labelformat=empty}
        \addtocounter{subfigure}{-1}
    }
    
    \caption{\small t-SNE plots of the features from the whole image classifier, indicating class and domain.}
    \label{fig:tsne_features_whole_image}
\end{figure}

Figure \ref{fig:tsne_features_whole_image} shows the features t-SNE plot for the whole image classifier. It can be seen that the features of the CE model can be easily separated by domain, despite the features of the CE patch-classifier being domain-invariant for most classes (we recall that the whole image classifier was obtained by extending the patch-classifier, pre-trained on CBIS-DDSM). We hypothesize that this is caused by the maladaptation of the normal patches for the CE model in Figure \ref{fig:features_patch_clsf_ddsm}, as every mammography image contains many normal regions. On the other hand, the features of the SupContr+LCP and SupContr+CE models are domain-invariant.

\section{Conclusions}

In this work, we mathematically showed that minimizing two standard contrastive losses - NT-Xent loss and Supervised Contrastive loss - decreases the CMMD and thus performs Domain Adaptation. Moreover, it improves class-separability in the feature space, which is often associated to higher downstream task performance. These findings offer a theoretical foundation for the growing adoption of Contrastive Learning as an effective approach for Domain Adaptation. Our theoretical results were further validated through numerical experiments, which demonstrated that minimizing the Supervised Contrastive Loss consistently improved Domain Adaptation and class separability, leading to enhanced classification performance in most cases. Considering these theoretical and empirical results, we conclude that Contrastive Learning can be effectively used for attaining Domain Adaptation while maintaining or improving class-separability in the feature space. Future research should explore the boundaries of these improvements in classification performance, particularly regarding the impact of weight initialization and the role of Transfer Learning.

\bibliography{contr_ler_da, dl_mammo_classif, domain_adaptation, federated_learning_medical_img, general_deep_learning, semi_sup_learning, new_bib, databases, clinical_mammography}
\bibliographystyle{icml2024}

\newpage
\appendix
\onecolumn

\section{Numerical experiments details}
\label{app:exp_details}

For each dataset, all the models were trained during the same number of epochs, which was set by making sure that all the models had converged. The patch-classifiers with clinical patches were trained for 250 epochs, the patch-classifiers with synthetic patches for 100 epochs, and the whole image classifiers for 200 epochs. The models with the best performance on the validation set (AUC for binary classifiers and AUC OvO for multi-class classifiers) were retained after each training round.

All models were optimized using Stochastic Gradient Descent (SGD) without momentum, with the learning rate scheduled by a cosine annealing scheduler \cite{cosine_annealing} with period $T=4$ epochs. The base learning rate was set to $10^{-4}$ for clinical images and patches, and to $10^{-3}$ for synthetic patches. To achieve balanced batches, less-represented classes, such as malignant ones, were oversampled. The batch size was set to 8 for the whole image classifier (the maximum value that fit into the GPU RAM) and 30 for the patch classifiers. All models used a weight decay of $10^{-4}$ and did not employ dropout. Data augmentation for patch classifiers included vertical and horizontal flips, as well as rotations by $90^\circ$, $180^\circ$, and $270^\circ$. For the whole image classifier, only horizontal flips were used. All images were used at their original resolutions, as resizing complicates the detection of small lesions \cite{Quintana2023}. For training the whole image classifier, the first three dense blocks of the DenseNet backbone were frozen to reduce GPU RAM usage, which did not negatively impact performance.

For training with the Supervised Contrastive loss, the temperature was set to $\tau = 0.5$ for synthetic patches. For clinical patches and full images, it was maintained at $\tau = 0.5$ for the first 50 epochs, then linearly decreased over the next 100 epochs to $\tau = 0.1$, where it remained constant for the final epochs (100 epochs for the patch-classifier and 50 for the whole image classifier). When fine-tuning linear layers, i.e., LCP, 20 epochs were used.

Each training was conducted on a 24 GB Nvidia Quadro RTX 6000 GPU. Training each patch-classifier with clinical patches took approximately 3 days, while training each whole image classifier took about 2.5 days.

\section{Parameters for generating the synthetic patch dataset}
\label{app:synthetic_params}

The patch size is set to $256 \times 256$ pixels, and $\beta$ varies between 1.2 and 1.6 (see Equation \eqref{eq:low_pass_filter}). Calcifications are organized into clusters containing 5 to 12 instances within square areas with side lengths ranging from 15 to 60 pixels. Calcification intensity spans 90\% to 100\% of the maximum image intensity. Masses are represented as Gaussian-shaped profiles with radii ranging from 5 to 45 pixels. The radii may differ along the two axes, resulting in oval or circular shapes. The intensity at the center of each mass is between 90\% and 100\% of the maximum image intensity. All parameters are adjustable using the code provided in the supplementary material.

\section{Proof of Lemma \ref{lemma:sup_contr_and_cmmd}}
\label{app:proof-lemma-contr-cmmd}

\begin{proof}

Starting with the definition of the Contrastive loss from Lemma \ref{lemma:contr_loss_with_expectations}, we seek to linearize the first term of Equation \eqref{eq:contr_expectations}.

By using the 2-nd order Taylor development of the exponential around $ \mathbb{E}_{X'  \sim \pi_p}[k(X, X') / \tau] $, we can write
\begin{dmath*}
    e^{k(X, X') / \tau} \approx \\
    e^{\mathbb{E}_{X'  \sim \pi_p}[ k(X, X') / \tau]} \left( 1 + \frac{1}{\tau} k(X, X') - \frac{1}{\tau} \mathbb{E}_{X'  \sim \pi_p}[ k(X, X') ] \\
    + \frac{1}{2\tau^2} ( k(X, X') - \mathbb{E}_{X'  \sim \pi_p}[ k(X, X') ] )^2  + \mathcal{O} \left( \frac{(k(X,X') - \mathbb{E}_{X'  \sim \pi_p}[ k(X, X') ] )^3}{\tau^3} \right) \right).
\end{dmath*}
Then, we have
\begin{dmath*}
   \mathbb{E}_{X  \sim \pi_p} \left[ \log \mathbb{E}_{X'  \sim \pi_p} \left[ e^{k(X, X') / \tau} \right] \right] =  \\ \mathbb{E}_{X  \sim \pi_p} \left[ \log \mathbb{E}_{X'  \sim \pi_p} \left[ e^{\mathbb{E}_{X'  \sim \pi_p}[ k(X, X') ] / \tau} \left( 1 +
   \frac{1}{\tau} k(X, X') - \frac{1}{\tau} \mathbb{E}_{X'  \sim \pi_p}[ k(X, X') ] + \\ \frac{1}{2\tau^2}( k(X, X') - \mathbb{E}_{X'  \sim \pi_p}[ k(X, X') ] )^2  + \mathcal{O} \left( \frac{(k(X,X') - \mathbb{E}_{X'  \sim \pi_p}[ k(X, X') ] )^3}{\tau^3} \right) \right) \right] \right],
\end{dmath*}

\begin{dmath*}
   \mathbb{E}_{X  \sim \pi_p} \left[ \log \mathbb{E}_{X'  \sim \pi_p} \left[ e^{k(X, X') / \tau} \right] \right] = \\ 
   \mathbb{E}_{X  \sim \pi_p} \left[ \log \left( e^{\mathbb{E}_{X'  \sim \pi_p} \left[ k(X, X') \right] / \tau } \left( 1 + \frac{1}{\tau}  \mathbb{E}_{X'  \sim \pi_p} [k(X, X')] - \frac{1}{\tau} \mathbb{E}_{X'  \sim \pi_p}[ k(X, X') ] + \\ \mathbb{E}_{X'  \sim \pi_p} \left[ \frac{1}{2\tau^2} \left( k(X, X') - \mathbb{E}_{X'  \sim \pi_p} [ k(X, X') ] \right)^2 \right]  + \mathbb{E}_{X'  \sim \pi_p} \left[ \mathcal{O} \left( \frac{(k(X,X') - \mathbb{E}_{X'  \sim \pi_p}[ k(X, X') ] )^3}{\tau^3} \right) \right] \right) \right) \right],
\end{dmath*}

\begin{dmath*}
   \mathbb{E}_{X  \sim \pi_p} \left[ \log \mathbb{E}_{X'  \sim \pi_p} \left[ e^{ k(X, X') / \tau} \right] \right] = \\
   \mathbb{E}_{X  \sim \pi_p} \left[ \log \left( e^{\mathbb{E}_{X'  \sim \pi_p} \left[ k(X, X') / \tau \right] } \left( 1 + \mathbb{E}_{X'  \sim \pi_p} \left[ \frac{1}{2\tau^2} \left( k(X, X' ) - \mathbb{E}_{X'  \sim \pi_p} \left[ k(X, X') \right] \right)^2 \right] +
   \\
   \mathcal{O} \left( \frac{ \mathbb{E}_{X'  \sim \pi_p} \left[ (k(X,X') - \mathbb{E}_{X'  \sim \pi_p}[ k(X, X') ] )^3 \right] }{\tau^3} \right) \right) \right) \right],
\end{dmath*}

\begin{dmath}
   \mathbb{E}_{X  \sim \pi_p} \left[ \log \mathbb{E}_{X'  \sim \pi_p} \left[ e^{k(X, X') / \tau} \right] \right] =  
   \\ 
   \mathbb{E}_{X  \sim \pi_p} \left[ \log \left( e^{\mathbb{E}_{X'  \sim \pi_p} \left[ k(X, X') \right] / \tau} \left( 1 + \mathcal{O} \left( \frac{ \mathbb{E}_{X'  \sim \pi_p} \left[ (k(X,X') - \mathbb{E}_{X'  \sim \pi_p}[ k(X, X') ] )^3 \right] }{\tau^3} \right) \right) 
   \\
   + e^{\mathbb{E}_{X'  \sim \pi_p} \left[ k(X, X') \right] / \tau} \frac{1}{2\tau^2} \mathbb{E}_{X'  \sim \pi_p} \left[ ( k(X, X') - 
   \mathbb{E}_{X'  \sim \pi_p} \left[ k(X, X') \right] )^2  \right] \right) \right].
\label{eq:log_exp_afer_exp_taylor}
\end{dmath}
They 1-st order Taylor expansion of $f(x) = \log (a +b x) $ around zero is given by:
\begin{equation*}
    \log (a + b x) \approx  \log a + \frac{b}{a} x + \mathcal{O} \left( \frac{b^2}{a^2} x^2 \right).
\end{equation*}
In Equation \eqref{eq:log_exp_afer_exp_taylor}, we have:
\begin{equation*}
    \begin{cases}
    a = e^{\mathbb{E}_{X'  \sim \pi_p} \left[ k(X, X') \right] / \tau} \left( 1 + \mathcal{O} \left( \frac{ \mathbb{E}_{X'  \sim \pi_p} \left[ (k(X,X') - \mathbb{E}_{X'  \sim \pi_p}[ k(X, X') ] )^3 \right] }{\tau^3} \right) \right) \\
    b = \frac{1}{2} e^{\mathbb{E}_{X'  \sim \pi_p} \left[ k(X, X') \right] / \tau} \\
    x = \mathbb{E}_{X'  \sim \pi_p} \left[ \frac{( k(X, X') - 
   \mathbb{E}_{X'  \sim \pi_p} \left[ k(X, X') \right] )^2}{\tau^2} \right],
    \end{cases}
\end{equation*}
which implies
\begin{dmath*}
    \log a 
    = \log \left( e^{\mathbb{E}_{X'  \sim \pi_p} \left[ k(X, X') \right] / \tau} \left( 1 + \mathcal{O} \left( \frac{ \mathbb{E}_{X'  \sim \pi_p} \left[ (k(X,X') - \mathbb{E}_{X'  \sim \pi_p}[ k(X, X') ] )^3 \right] }{\tau^3} \right) \right)  \right)
    = \frac{1}{\tau} \mathbb{E}_{X'  \sim \pi_p} \left[ k(X, X') \right] + \mathcal{O} \left( \log \left( 1 + \frac{ \mathbb{E}_{X'  \sim \pi_p} \left[ (k(X,X') - \mathbb{E}_{X'  \sim \pi_p}[ k(X, X') ] )^3 \right] }{\tau^3} \right) \right),
\end{dmath*}
and
\begin{dmath*}
    \frac{b}{a} = \frac{1}{2} \frac{1}{ 1 + \mathcal{O} \left( \frac{ \mathbb{E}_{X'  \sim \pi_p} \left[ (k(X,X') - \mathbb{E}_{X'  \sim \pi_p}[ k(X, X') ] )^3 \right] }{\tau^3} \right)}.
\end{dmath*}
We thus have
\begin{dmath*}
\mathbb{E}_{X  \sim \pi_p} \left[ \log \mathbb{E}_{X'  \sim \pi_p} \left[ e^{k(X, X') / \tau} \right] \right] \approx 
\\
\mathbb{E}_{X  \sim \pi_p} \left[ \frac{1}{\tau} \mathbb{E}_{X'  \sim \pi_p} \left[ k(X, X') \right] + \mathcal{O} \left( \log \left( 1 + \frac{ \mathbb{E}_{X'  \sim \pi_p} \left[ (k(X,X') - \mathbb{E}_{X'  \sim \pi_p}[ k(X, X') ] )^3 \right] }{\tau^3} \right) \right)
+ 
\frac{1}{2} \frac{1}{ 1 + \mathcal{O} \left( \frac{ \mathbb{E}_{X'  \sim \pi_p} \left[ (k(X,X') - \mathbb{E}_{X'  \sim \pi_p}[ k(X, X') ] )^3 \right] }{\tau^3} \right)} \mathbb{E}_{X'  \sim \pi_p} \left[ \frac{( k(X, X') - 
\mathbb{E}_{X'  \sim \pi_p} \left[ k(X, X') \right] )^2}{\tau^2} \right]
+
\\
\mathcal{O} \left(  \frac{1}{ \left( 1 + \mathcal{O} \left( \frac{ \mathbb{E}_{X'  \sim \pi_p} \left[ (k(X,X') - \mathbb{E}_{X'  \sim \pi_p}[ k(X, X') ] )^3 \right] }{\tau^3} \right) \right)^2 } \frac{\left( \mathbb{E}_{X'  \sim \pi_p} \left[ ( k(X, X') - \mathbb{E}_{X'  \sim \pi_p} \left[ k(X, X') \right] )^2  \right]  \right)^2}{\tau^4} \right) \right].
\end{dmath*}
However, in the vicinity of $ \mathbb{E}_{X'  \sim \pi_p} \left[ k(X, X') \right] $, where the Taylor approximation is valid, we have that \\ $ | \mathbb{E}_{X'  \sim \pi_p} \left[ k(X,X') - \mathbb{E}_{X'  \sim \pi_p} \left[ k(X, X') \right] \right] | < \tau $ and
\begin{dmath*}
\mathbb{E}_{X  \sim \pi_p} \left[ \log \mathbb{E}_{X'  \sim \pi_p} \left[ e^{k(X, X') / \tau} \right] \right] 
\\ \approx 
\mathbb{E}_{X  \sim \pi_p} \left[ \frac{1}{\tau} \mathbb{E}_{X'  \sim \pi_p} \left[ k(X, X') \right]
+ 
\frac{1}{2} \mathbb{E}_{X'  \sim \pi_p} \left[ \frac{( k(X, X') - 
\mathbb{E}_{X'  \sim \pi_p} \left[ k(X, X') \right] )^2}{\tau^2} \right]
+
\\
\mathcal{O} \left( \frac{\left( \mathbb{E}_{X'  \sim \pi_p} \left[ ( k(X, X') - \mathbb{E}_{X'  \sim \pi_p} \left[ k(X, X') \right] )^2  \right]  \right)^2}{\tau^4} \right) \right]
\\ \approx
\frac{1}{\tau} \mathbb{E}_{X, X'  \sim \pi_p} \left[ k(X, X') \right]
+ 
\frac{1}{2 \tau^2} \mathbb{E}_{X  \sim \pi_p} \left[ \mathbb{E}_{X'  \sim \pi_p} \left[( k(X, X') - 
\mathbb{E}_{X'  \sim \pi_p} \left[ k(X, X') \right] )^2 \right] \right]
+
\\
\mathcal{O} \left( \frac{\mathbb{E}_{X  \sim \pi_p} \left[ \left( \mathbb{E}_{X'  \sim \pi_p} \left[ ( k(X, X') - \mathbb{E}_{X'  \sim \pi_p} \left[ k(X, X') \right] )^2  \right] \right)^2 \right]}{\tau^4} \right) 
\\ \approx
\frac{1}{\tau} \mathbb{E}_{X, X'  \sim \pi_p} \left[ k(X, X') \right]
+ 
\frac{1}{2 \tau^2} \mathbb{E}_{X  \sim \pi_p} \left[ {\mathrm{Var}}_{X'  \sim \pi_p} \left[ k(X, X') \right] \right]
+
\\
\mathcal{O} \left( \frac{\mathbb{E}_{X  \sim \pi_p} \left[ {\mathrm{Var}}_{X'  \sim \pi_p} \left[ k(X, X') \right]^2 \right]}{\tau^4} \right),
\end{dmath*}
and Equation \eqref{eq:contr_expectations} can then be re-written as
\begin{dmath*}
    \label{eq:contr_expectation_rewritten}
    \mathcal{L}_{Contr} \approx \frac{1}{\tau} \mathbb{E}_{X, X'  \sim \pi_p} \left[ k(X, X') \right] + \frac{1}{2\tau^2} \mathbb{E}_{X \sim \pi_p} \left[ {\mathrm{Var}}_{X'  \sim \pi_p}  \left[ k(X, X') \right] \right] - \frac{1}{\tau} \mathbb{E}_{X,X'\sim Pos } [ k(X,X') ]    + \log \left( | \mathcal{B} | -1 \right) + \mathcal{O} \left( \frac{\mathbb{E}_{X  \sim \pi_p} \left[ {\mathrm{Var}}_{X'  \sim \pi_p} \left[ k(X, X') \right]^2 \right]}{\tau^4} \right).
\end{dmath*}
By solving for $ \mathbb{E}_{X,X'\sim Pos } [ k(X,X') ] $, 
\begin{dmath}
\label{eq:contr_loss_expectation_arbitrary_p}
    \mathbb{E}_{X,X'\sim Pos } [ k(X,X') ] \approx  \mathbb{E}_{X, X'  \sim \pi_p} \left[ k(X, X') \right] + \frac{1}{2\tau} \mathbb{E}_{X \sim \pi_p} \left[ {\mathrm{Var}}_{X'  \sim \pi_p}  \left[ k(X, X') \right] \right] + \tau \log \left( | \mathcal{B} | -1 \right) - \tau \mathcal{L}_{Contr} + \mathcal{O} \left( \frac{\mathbb{E}_{X  \sim \pi_p} \left[ {\mathrm{Var}}_{X'  \sim \pi_p} \left[ k(X, X') \right]^2 \right]}{\tau^3} \right).
\end{dmath}
As in the CMMD the positive pairs are sampled from a mixture distribution with equiprobable domains, we need to set $p=1/2$ in Equation \eqref{eq:contr_loss_expectation_arbitrary_p}. This means that if the observed mixture distribution does not have equiprobable domains, the batch size has to be artificially constructed to have them. By setting $p=1/2$ and using Equation \eqref{eq:contr_loss_expectation_arbitrary_p} with Lemma \ref{lemma:cmmd_with_expactations}, we have
\begin{dmath*}
    CMMD^2(\mathcal{D}_0, \mathcal{D}_1) \approx 2  \mathbb{E}_{C \sim Q} \left[ \mathbb{E}_{X,X' \sim \pi^{\mathcal{X} | \mathcal{Y}}_{0,C}} \left[ k(X,X') \right] + \mathbb{E}_{X, X' \sim \pi^{\mathcal{X} | \mathcal{Y}}_{1,C}} \left[ k(X,X') \right] \right] \\ - 4 \left( \mathbb{E}_{X, X'  \sim \pi^{\mathcal{X}}_{0.5}} \left[ k(X, X') \right] + \frac{1}{2\tau} \mathbb{E}_{X \sim \pi^{\mathcal{X}}_{0.5}} \left[ {\mathrm{Var}}_{X'  \sim \pi^{\mathcal{X}}_{0.5}}  \left[ k(X, X') \right] \right] + \tau \log \left( | \mathcal{B} | -1 \right) \\ - \tau \mathcal{L}_{Contr} + \mathcal{O} \left( \frac{\mathbb{E}_{X  \sim \pi^{\mathcal{X}}_{0.5}} \left[ {\mathrm{Var}}_{X'  \sim \pi^{\mathcal{X}}_{0.5}} \left[ k(X, X') \right]^2 \right]}{\tau^3} \right) \right),
\end{dmath*}
and by re-organizing
\begin{dmath*}
    \mathcal{L}_{Contr} 
    \approx
    \frac{1}{4\tau} CMMD^2(\mathcal{D}_0, \mathcal{D}_1) + \frac{1}{\tau} \mathbb{E}_{X, X'  \sim \pi^{\mathcal{X}}_{0.5}} \left[ k(X, X') \right] - \frac{1}{2\tau} \mathbb{E}_{C \sim Q} \left[ \mathbb{E}_{X,X' \sim \pi^{\mathcal{X} | \mathcal{Y}}_{0,C}} \left[ k(X,X') \right] + \mathbb{E}_{X, X' \sim \pi^{\mathcal{X} | \mathcal{Y}}_{1,C}} \left[ k(X,X') \right] \right] + \frac{1}{2\tau^2} \mathbb{E}_{X \sim \pi^{\mathcal{X}}_{0.5}} \left[ \mathrm{Var}_{X'  \sim \pi^{\mathcal{X}}_{0.5}}  \left[ k(X, X') \right] \right] + \log \left( | \mathcal{B} | -1 \right) \\ + \mathcal{O} \left( \frac{\mathbb{E}_{X  \sim \pi^{\mathcal{X}}_{0.5}} \left[ {\mathrm{Var}}_{X'  \sim \pi^{\mathcal{X}}_{0.5}} \left[ k(X, X') \right]^2 \right]}{\tau^4} \right).
\end{dmath*}
Multiplying the two sides by $\tau$ proves the relationship.

\end{proof}

\section{Proof of Lemma \ref{lemma:contr_loss_hsic_mmd_bound}}
\label{app:proof_contr_loss_hsic_mmd_bound}

Lemma \ref{lemma:contr_loss_hsic_mmd_bound} has been proven by \citet{li2021selfsupervised} for the NT-Xent loss. Here, we prove it is also valid for the Supervised Contrastive loss.

\begin{proof}

Using Lemmas \ref{lemma:contr_loss_with_expectations} and \ref{lemma:hsic_sup_contrastive_loss} to write the contrastive losses (Supervised and NT-Xent) and $\text{HSIC}(X,Y)$ in terms of the expectations, and applying Theorem B.1 of \citet{li2021selfsupervised} we obtain:

\begin{equation}
    \label{eq:contr_loss_hsic_bound}
    - \text{HSIC}(X,Y) + \gamma \text{HSIC}(X,X) + \mathcal{O} \left( \mathrm{Var} \left[ k \left( X, X' \right) \right] \right) \leqslant \mathcal{L}_{Contr}.
\end{equation}
Theorem B.1 gives the conditions for the kernels and for $\gamma$. Finally, in the Appendix B, \citet{li2021selfsupervised} prove the following relationship between the inter-class MMD and HSIC(X,Y):

\begin{equation}
    \label{eq:mmd_and_hsic}
    \underbrace{\mathbb{E}_{C_1, C_2 \sim \pi^{\mathcal{Y}}_{0.5}} \left[ \| \mathbb{E}_{X \sim \pi^{\mathcal{X}|\mathcal{Y}}_{0.5, C_1}} [ \phi(X) ] - \mathbb{E}_{X \sim \pi^{\mathcal{X}|\mathcal{Y}}_{0.5, C_2}} [ \phi(X) ] \|^2\right]}_{\text{inter-class MMD}} = \alpha \ \text{HSIC}(X,Y),
\end{equation}
where $ \alpha $ is a proportionality constant which depends on problem parameters, such as the number of classes and the kernels used. Combining Equations \eqref{eq:mmd_and_hsic} and \eqref{eq:contr_loss_hsic_bound} proves the lemma.

\end{proof}

\section{Useful lemmas}
\label{app:useful_lemmas}

\begin{lemma}
\label{lemma:kernel_over_y_simplified}

Let $l(y,y')=\langle y, y' \rangle_{\mathcal{Y}}$ be a kernel over $\mathcal{Y}$, which is assumed to be a function of $y \cdot y'$ or $ \| y - y' \|$. Then, it can be written as:
\begin{equation}
    l(y, y') = 
        \begin{cases}
            l_1 & \quad \text{if} \quad y = y' \\
            l_0 & \quad \text{otherwise}
        \end{cases}
    \: =  \: \Delta l \ \one_{\{ y = y' \} } + l_0,
\end{equation}
where $ \Delta = l_1 - l_0 $ and $\one_{\{ . \} }$ is the indicator function.
    
\end{lemma}

\begin{proof}

Assuming that the label of each sample $y$ is in one-hot format, any kernel that is a function of $y \cdot y'$ or $ \| y - y'\| $ can take only two possible values: $l_1$ when the two data points share the same label, i.e., $y = y'$, and $l_0$ when they have different labels, i.e., $y \neq y'$.
    
\end{proof}

\begin{lemma}
\label{lemma:expectation_mixture_of_domains}

Given two domains $\mathcal{D}_0 = \{ \mathcal{X} \times \mathcal{Y}, \pi_0 \}$ and $\mathcal{D}_1 = \{ \mathcal{X} \times \mathcal{Y} \times \mathcal{Y}, \pi_1 \}$. The expectation of an arbitrary integrable function $ g: \mathcal{X} \times \mathcal{X} \rightarrow \mathbb{R} $ on the mixture domain $ \mathcal{D}_p = \{ \mathcal{X} \times \mathcal{Y}, \pi_p \} $ is given by:
\begin{dmath}
    \mathbb{E}_{X,X' \sim \pi^{\mathcal{X}}_p} \left[ g(X,X') \right] 
    = 
    p^2 \mathbb{E}_{X,X' \sim \pi^{\mathcal{X}}_1} \left[ g(X,X') \right] + p (1-p) \mathbb{E}_{X \sim \pi^{\mathcal{X}}_1, X' \sim \pi^{\mathcal{X}}_0} \left[ g(X,X') \right] + p (1-p) \mathbb{E}_{X \sim \pi^{\mathcal{X}}_0, X' \sim \pi^{\mathcal{X}}_1} \left[ g(X,X') \right] + (1-p)^2 \mathbb{E}_{X,X' \sim \pi^{\mathcal{X}}_0} \left[ g(X,X') \right].
\end{dmath}
If the variables $X$ and $X'$ are interchangeable in $ g(X,X')$, then the expectation is given by:
\begin{dmath}
    \mathbb{E}_{X,X' \sim \pi^{\mathcal{X}}_p} \left[ g(X,X') \right] 
    = 
    p^2 \mathbb{E}_{X,X' \sim \pi^{\mathcal{X}}_1} \left[ g(X,X') \right] + 2 p (1-p) \mathbb{E}_{X \sim \pi^{\mathcal{X}}_1, X' \sim \pi^{\mathcal{X}}_0} \left[ g(X,X') \right] + (1-p)^2 \mathbb{E}_{X,X' \sim \pi^{\mathcal{X}}_0} \left[ g(X,X') \right].
\end{dmath}

\end{lemma}

\begin{proof}
By defining a binary hidden variable $Z \sim Ber(p)$ that determines the original distribution from which $X$ is sampled ($ Z \in \{ 1, 2 \}$), we have:
\begin{dmath*}
\mathbb{E}_{X,X' \sim \pi^{\mathcal{X}}_p} \left[ g(X,X') \right]
     = \mathbb{E}_{Z,Z' \sim Ber(p)} \left[ \mathbb{E}_{X,X' \sim \pi^{\mathcal{X}}_p} \left[ g(X,X') | {Z=z, Z'=z'} \right] \right] 
    \\ = 
    p^2 \mathbb{E}_{X,X' \sim \pi^{\mathcal{X}}_p} \left[ g(X,X') | {Z=1, Z'=1} \right] + p (1-p)\mathbb{E}_{X,X' \sim \pi^{\mathcal{X}}_p} \left[ g(X,X') | {Z=1, Z'=2} \right] + (1-p)p \mathbb{E}_{X,X' \sim \pi^{\mathcal{X}}_p} \left[ g(X,X') | {Z=2, Z'=1} \right] \\ \hspace*{0.8em} + (1-p)^2 \mathbb{E}_{X,X' \sim \pi^{\mathcal{X}}_p} \left[ g(X,X') | {Z=2, Z'=2} \right]
    = 
    p^2 \mathbb{E}_{X,X' \sim \pi^{\mathcal{X}}_1} \left[ g(X,X') \right] + p (1-p) \mathbb{E}_{X \sim \pi^{\mathcal{X}}_1, X' \sim \pi^{\mathcal{X}}_0} \left[ g(X,X') \right] + (1-p)p \mathbb{E}_{X \sim \pi^{\mathcal{X}}_0, X' \sim \pi^{\mathcal{X}}_1} \left[ g(X,X') \right] \\ \hspace*{0.8em} + (1-p)^2 \mathbb{E}_{X,X' \sim \pi^{\mathcal{X}}_0} \left[ g(X,X') \right].
\end{dmath*}
If $g(X,X')=g(X',X)$ we have that $ \mathbb{E}_{X \sim \pi^{\mathcal{X}}_1, X' \sim \pi^{\mathcal{X}}_0} \left[ g(X,X') \right] = \mathbb{E}_{X \sim \pi^{\mathcal{X}}_0, X' \sim \pi^{\mathcal{X}}_1} \left[ g(X,X') \right] $ and thus
\begin{dmath*}
\mathbb{E}_{X,X' \sim \pi^{\mathcal{X}}_p} \left[ g(X,X') \right] 
    = 
    p^2 \mathbb{E}_{X,X' \sim \pi^{\mathcal{X}}_1} \left[ g(X,X') \right] + 2 p (1-p) \mathbb{E}_{X \sim \pi^{\mathcal{X}}_1, X' \sim \pi^{\mathcal{X}}_0} \left[ g(X,X') \right] + (1-p)^2 \mathbb{E}_{X,X' \sim \pi^{\mathcal{X}}_0} \left[ g(X,X') \right].
\end{dmath*}
    
\end{proof}

\begin{lemma}
\label{lemma:contr_loss_with_expectations}

The NT-Xent loss and the Supervised Contrastive loss can be written in terms of the expectation by the following equation, with $ k(X,X') = \phi(X)^T \phi(X') = Z^T Z' $:
\begin{equation}
    \label{eq:contr_expectations}
    \mathcal{L}_{Contr} \approx \mathbb{E}_{X  \sim \pi^{\mathcal{X}}_p} \left[ \log \ \mathbb{E}_{X' \sim \pi^{\mathcal{X}}_p} \left[ e^{k(X,X') / \tau} \right] \right]
    - \frac{1}{\tau} \mathbb{E}_{X,X' \sim Pos} \left[ k(X,X') \right] + \log ( | \mathcal{B}| - 1 ),
\end{equation}
where $ \pi^{\mathcal{X}}_p $ is the probability measure of the mixture of the two domains, with a mixture probability $p$, and where $\mathcal{L}_{Contr}$ represents any of the two Contrastive losses.

\end{lemma}

\begin{proof}[Proof (NT-Xent loss).]

From Equation \eqref{eq:def_self_sup_contr}, and by replacing $ k(X,X') = \phi(X)^T \phi(X') = Z^T Z'$
\begin{equation*}
    \mathcal{L}_{Contr} = \frac{1}{| \mathcal{B} |} \sum_{x_i \in  \mathcal{B}} \log \sum_{l \in \mathcal{A}(i)} e^{k(x_i, x_l) / \tau} - \frac{1}{| \mathcal{B} | \tau} \sum_{x_i \in  \mathcal{B}} k(x_i, x_{j(i)}).
\end{equation*}
By denoting as $\hat{\mathbb{E}}$ the estimation of the expectation, we can write
\begin{equation}
\label{eq:contr_loss_expectations_intermediary_eq}
    \mathcal{L}_{Contr} = \hat{\mathbb{E}}_{X \sim \pi^{\mathcal{X}}_p} \left[ \log \left( ( | \mathcal{B}| - 1 ) \hat{\mathbb{E}}_{X' \sim \pi^{\mathcal{X}}_p, X' \neq X} [ e^{k(X,X') / \tau} ] \right) \right]
    - \frac{1}{\tau} \hat{\mathbb{E}}_{X,X' \sim Pos} \left[ k(X,X') \right].
\end{equation}
By using the product property of the logarithmic we get
\begin{equation*}
    \mathcal{L}_{Contr} = \hat{\mathbb{E}}_{X  \sim \pi^{\mathcal{X}}_p} \left[ \log \ \hat{\mathbb{E}}_{X' \sim \pi^{\mathcal{X}}_p, X' \neq X} \left[ e^{k(X,X') / \tau} \right] \right]
    - \frac{1}{\tau} \hat{\mathbb{E}}_{X,X' \sim Pos} \left[ k(X,X') \right] + \log ( | \mathcal{B}| - 1 ).
\end{equation*}
For $ | \mathcal{B} | $ sufficiently large, $ \hat{\mathbb{E}}_{X \sim \pi^{\mathcal{X}}_p} [ ... ] \approx \mathbb{E}_{X \sim \pi^{\mathcal{X}}_p} [ ... ] $,  $ \hat{\mathbb{E}}_{X, X' \sim Pos} [ ... ] \approx \mathbb{E}_{X,X' \sim Pos} [ ... ] $ and \\ $ \hat{\mathbb{E}}_{X' \sim \pi^{\mathcal{X}}_p, X \neq X'} [ ... ] \approx \mathbb{E}_{X' \sim \pi^{\mathcal{X}}_p X \neq X' } [ ... ] = \mathbb{E}_{X' \sim \pi^{\mathcal{X}}_p} [ ... ] $, as $ \pi^{\mathcal{X}}_p (X=X') = 0 \quad \forall X,X' \in \mathcal{X} \times \mathcal{X}  $. Thus,
\begin{equation*}
    \mathcal{L}_{Contr} \approx \mathbb{E}_{X  \sim \pi^{\mathcal{X}}_p} \left[ \log \ \mathbb{E}_{X' \sim \pi^{\mathcal{X}}_p} \left[ e^{k(X,X') / \tau} \right] \right]
    - \frac{1}{\tau} \mathbb{E}_{X,X' \sim Pos} \left[ k(X,X') \right] + \log ( | \mathcal{B}| - 1 ).
\end{equation*}
 
\end{proof}

\begin{proof}[Proof (Supervised Contrastive loss).]

From Equation \eqref{eq:def_sup_contr}, and by replacing $ k(X,X') = \phi(X)^T \phi(X') = Z^T Z'$
\begin{equation*}
    \mathcal{L}_{Contr} = \frac{1}{| \mathcal{B} |} \sum_{i \in | \mathcal{B} | } \frac{1}{| \mathcal{P}(i) |} \sum_{j \in \mathcal{P}(i)} \log {\sum_{l \in \mathcal{A}(i)} e^{k(x_i,x_l) / \tau}} 
    - \frac{1}{| \mathcal{B} | \tau} \sum_{i \in | \mathcal{B} | } \frac{1}{| \mathcal{P}(i) |} \sum_{j \in \mathcal{P}(i)} k(x_i,x_j).
\end{equation*}
By rewriting in terms of the expectations, we obtain
\begin{equation*}
    \mathcal{L}_{Contr} = \hat{\mathbb{E}}_{X,X' \sim Pos} \left[ \log \left( ( | \mathcal{B}| - 1 ) \hat{\mathbb{E}}_{X'' \sim \pi^{\mathcal{X}}_p, X'' \neq X} [ e^{k(X,X'') / \tau} ] \right) \right]
    - \frac{1}{\tau} \hat{\mathbb{E}}_{X,X' \sim Pos} \left[ k(X,X') \right].
\end{equation*}
As the first term does not depend on $X'$, we can rewrite
\begin{equation*}
    \mathcal{L}_{Contr} = \hat{\mathbb{E}}_{X \sim \pi^{\mathcal{X}}_p} \left[ \log \left( ( | \mathcal{B}| - 1 ) \hat{\mathbb{E}}_{X' \sim \pi^{\mathcal{X}}_p, X' \neq X} [ e^{k(X,X') / \tau} ] \right) \right]
    - \frac{1}{\tau} \hat{\mathbb{E}}_{X,X' \sim Pos} \left[ k(X,X') \right],
\end{equation*}
where we have renamed $X''$ as $X'$, and have assumed a balanced batch in terms of the classes. This is Equation \eqref{eq:contr_loss_expectations_intermediary_eq}, and we can then proceed as in the Self-supervised Learning case to obtain Equation \eqref{eq:contr_expectations}.

\end{proof}

\begin{lemma}
\label{lemma:cmmd_with_expactations}

By considering a mapping of the type $ \phi: \mathcal{X} \rightarrow \mathcal{Z} \subseteq \mathbb{R}^m $, the square of the CMMD can be written in terms of the expectation by the following equation:
\begin{dmath}
    \label{eq:cmmd_def_expectations}
    \text{CMMD}^2(\mathcal{D}_0, \mathcal{D}_1, \phi)
    = 
    2 \mathbb{E}_{C \sim \pi^{\mathcal{Y}}} \left[ \mathbb{E}_{X,X' \sim \pi^{\mathcal{X}|\mathcal{Y}}_{0, C}} \left[ k(X,X') \right] + \mathbb{E}_{X,X' \sim \pi^{\mathcal{X}|\mathcal{Y}}_{1, C}} \left[ k(X,X') \right] \right] 
    - 4 \mathbb{E}_{X,X'\sim Pos } \left[ k(X, X') | {0.5} \right],
\end{dmath}
where $ k(X,X') = \phi(X)^T \phi(X') $, $ \mathbb{E}_{C \sim \pi^{\mathcal{Y}}} \left[ \mathbb{E}_{X,X' \sim \pi^{\mathcal{X}|\mathcal{Y}}_{0, C}} \left[ k(X,X') \right] + \mathbb{E}_{X,X' \sim \pi^{\mathcal{X}|\mathcal{Y}}_{1, C}} \left[ k(X,X') \right] \right] $ is the mean similarity in each class $C \in \{ 1, ..., c \} $ and domain $D \in \{ 1, 2 \} $, $ X,X'\sim Pos $ indicates that $X$ and $X'$ are positive pairs (instances with the same class, and same or different domain), and $ p$ is the domain mixture probability. The fact that $p=1/2$ states that for the CMMD definition, the two domains are equiprobable.

\end{lemma}

\begin{proof}

As $ \phi: \mathcal{X} \rightarrow \mathcal{Z} \subseteq \mathbb{R}^m $, $ \langle \phi(X) , \phi(X') \rangle = \phi(X)^T \phi(X')  $ and by using the linearity of the expectation and inner product in Equation \eqref{eq:cmmd_def}, we obtain
\begin{dmath*}
    \text{CMMD}^2(\mathcal{D}_0, \mathcal{D}_1, \phi) \\ = 
    \mathbb{E}_{C \sim \pi^{\mathcal{Y}}} \left[ \langle \mathbb{E}_{X \sim \pi^{\mathcal{X}|\mathcal{Y}}_{0,C}} \left[ \phi(X) \right] - \mathbb{E}_{X \sim \pi^{\mathcal{X}|\mathcal{Y}}_{1,C}} \left[ \phi(X) \right], \mathbb{E}_{X \sim \pi^{\mathcal{X}|\mathcal{Y}}_{0,C}} \left[ \phi(X) \right] - \mathbb{E}_{X \sim \pi^{\mathcal{X}|\mathcal{Y}}_{1,C}} \left[ \phi(X) \right] \rangle \right]
    = 
    \mathbb{E}_{C \sim \pi^{\mathcal{Y}}} \left[  \langle \mathbb{E}_{X \sim \pi^{\mathcal{X}|\mathcal{Y}}_{0,C}} \left[ \phi(X) \right], \mathbb{E}_{X \sim \pi^{\mathcal{X}|\mathcal{Y}}_{0,C}} \left[ \phi(X) \right] \rangle -2 \langle \mathbb{E}_{X \sim \pi^{\mathcal{X}|\mathcal{Y}}_{0,C}} \left[ \phi(X) \right], \mathbb{E}_{X \sim \pi^{\mathcal{X}|\mathcal{Y}}_{1,C}} \left[ \phi(X) \right] \rangle + \langle \mathbb{E}_{X \sim \pi^{\mathcal{X}|\mathcal{Y}}_{1,C}} \left[ \phi(X) \right], \mathbb{E}_{X \sim \pi^{\mathcal{X}|\mathcal{Y}}_{1,C}} \left[ \phi(X) \right] \rangle  \right]
    = 
    \mathbb{E}_{C \sim \pi^{\mathcal{Y}}} \left[ \mathbb{E}_{X,X' \sim \pi^{\mathcal{X}|\mathcal{Y}}_{0, C}} \left[ \langle \phi(X) , \phi(X') \rangle \right] -2 \mathbb{E}_{X \sim \pi^{\mathcal{X}|\mathcal{Y}}_{1,C}, X' \sim \pi^{\mathcal{X}|\mathcal{Y}}_{1,C}} \left[ \langle \phi(X) , \phi(X') \rangle \right] + \mathbb{E}_{X,X' \sim \pi^{\mathcal{X}|\mathcal{Y}}_{1, C}} \left[ \langle \phi(X), \phi(X') \rangle \right]  \right]
    = 
    \mathbb{E}_{C \sim \pi^{\mathcal{Y}}} \left[ \mathbb{E}_{X,X' \sim \pi^{\mathcal{X}|\mathcal{Y}}_{0, C}} \left[ k(X,X') \right] -2 \mathbb{E}_{X \sim \pi^{\mathcal{X}|\mathcal{Y}}_{1,C}, X' \sim \pi^{\mathcal{X}|\mathcal{Y}}_{1,C}} \left[ k(X,X') \right] + \mathbb{E}_{X,X' \sim \pi^{\mathcal{X}|\mathcal{Y}}_{1, C}} \left[ k(X,X') \right]  \right],
\end{dmath*}
where it was used that $ k(x,y) = \langle \phi(x), \phi(y) \rangle $. The notation was simplified by dropping the explicit dependence on the space in the inner products and norms, i.e., $\langle ., . \rangle_{\mathcal{Z}} = \langle ., . \rangle $ and $ \| . \|_{\mathcal{Z}} = \| . \|$. By adding and subtracting the intra-domain similarities, we have
\begin{dmath*}
    \text{CMMD}^2(\mathcal{D}_0, \mathcal{D}_1, \phi)
    = 
    2 \mathbb{E}_{C \sim \pi^{\mathcal{Y}}} \left[ \mathbb{E}_{X,X' \sim \pi^{\mathcal{X}|\mathcal{Y}}_{0, C}} \left[ k(X,X') \right] + \mathbb{E}_{X,X' \sim \pi^{\mathcal{X}|\mathcal{Y}}_{1, C}} \left[ k(X,X') \right] \right] - \mathbb{E}_{C \sim \pi^{\mathcal{Y}}} \left[ \mathbb{E}_{X,X' \sim \pi^{\mathcal{X}|\mathcal{Y}}_{0, C}} \left[ k(X,X') \right] + 2 \mathbb{E}_{X \sim \pi^{\mathcal{X}|\mathcal{Y}}_{1,C}, X' \sim \pi^{\mathcal{X}|\mathcal{Y}}_{1,C}} \left[ k(X,X') \right] + \mathbb{E}_{X,X' \sim \pi^{\mathcal{X}|\mathcal{Y}}_{1, C}} \left[ k(X,X') \right]  \right].
\end{dmath*}
From Lemma \ref{lemma:expectation_mixture_of_domains}, if the mixture probability is $ p = 1/2 $, we have:
\begin{dmath}
\label{eq:cmmd_expectation_without_before_pos_pair_notation}
    \text{CMMD}^2(\mathcal{D}_0, \mathcal{D}_1, \phi) \smash{=} \\
    2 \mathbb{E}_{C \sim \pi^{\mathcal{Y}}} \left[ \mathbb{E}_{X,X' \sim \pi^{\mathcal{X}|\mathcal{Y}}_{0, C}} \left[ k(X,X') \right] + \mathbb{E}_{X,X' \sim \pi^{\mathcal{X}|\mathcal{Y}}_{1, C}} \left[ k(X,X') \right] \right] - 4 \mathbb{E}_{C \sim \pi^{\mathcal{Y}}} \left[  \mathbb{E}_{X,X' \sim \pi^{\mathcal{X}|\mathcal{Y}}_{m,0.5, C}} \left[ k(X,X') \right]  \right]
\end{dmath},
as the similarity is symmetric with respect to $X$ and $X'$. The first term of Equation \eqref{eq:cmmd_expectation_without_before_pos_pair_notation} contains the mean similarity inside each cluster (label and domain). The second term of Equation \eqref{eq:cmmd_expectation_without_before_pos_pair_notation} contains the mean similarity between features that share the same label, with different and same domain. 
In the Contrastive Learning literature, these are commonly denoted as positive pairs. Equation \eqref{eq:cmmd_expectation_without_before_pos_pair_notation} can then  be rewritten in terms of the expectation:
\begin{dmath*}
    \text{CMMD}^2(\mathcal{D}_0, \mathcal{D}_1, \phi)
    = 
    2 \mathbb{E}_{C \sim \pi^{\mathcal{Y}}} \left[ \mathbb{E}_{X,X' \sim \pi^{\mathcal{X}|\mathcal{Y}}_{0, C}} \left[ k(X,X') \right] + \mathbb{E}_{X,X' \sim \pi^{\mathcal{X}|\mathcal{Y}}_{1, C}} \left[ k(X,X') \right] \right] 
    - 4 \mathbb{E}_{X,X'\sim Pos } \left[ k(X, X') | {p = 1/2} \right],
\end{dmath*} 
where $p$ is the mixture probability.

\end{proof}

\begin{lemma}
\label{lemma:hsic_sup_contrastive_loss}

In a learning setting with $N$ data points sampled independently with the same probability, the $\text{HSIC}(X,Y)$ can be written as
\begin{equation}
    \label{eq:hsic_xy_sup}
    \text{HSIC}(X,Y) = \beta \ \mathbb{E}_{X,X' \sim Pos}\left[ k\left( X,X' \right) \right] - \beta \ \mathbb{E} \left[ k \left( X,X' \right) \right],
\end{equation}
where $X,X' \sim Pos$ means that the features sampled are positive pairs and $\beta$ is a constant. For the Supervised Contrastive loss $\beta = \frac{\Delta}{K}$, with $K$ the number of equiprobable classes and $\Delta $ a kernel-related constant, whereas for the Self-supervised Contrastive loss $\beta = \frac{\Delta}{N}$.
    
\end{lemma}

\begin{proof}[Proof (NT-Xent loss)]

Refer to Theorem A.1 of \citet{li2021selfsupervised}.

\end{proof}

\begin{proof}[Proof (Supervised Contrastive loss)]

We use Equation (2) of \citet{li2021selfsupervised} to write $\text{HSIC}(X,Y)$ as
\begin{equation}
    \label{eq:hsic_developed}
    \text{HSIC}(X,Y) = \mathbb{E} \left[ k(X,X') l(Y,Y') \right] - 2 \mathbb{E} \left[ k(X,X') l(Y,Y'') \right] + \mathbb{E} \left[ k(X,X') \right] \mathbb{E} \left[ l(Y,Y') \right].
\end{equation}
Using Lemma \ref{lemma:kernel_over_y_simplified}, the first term of Equation \eqref{eq:hsic_developed} yields:
\begin{dmath}
    \mathbb{E} \left[ k(X,X') l(Y,Y') \right] = \Delta l \ \mathbb{E} \left[ k(X,X') \one_{\{ Y = Y' \} } \right] + l_0 \mathbb{E} \left[ k(X,X') \right] \\
    = \Delta l \ \mathbb{E}_{Y,Y'} \left[ \mathbb{E}_{X,X'} \left[ k(X,X') \one_{\{ y = y' \} } | Y\smash{=}y, Y'\smash{=}y' \right] \right] + l_0 \mathbb{E} \left[ k(X,X') \right] \\
    = \Delta l \ \mathbb{P}(Y\smash{=}Y') \mathbb{E}_{X,X'} \left[ k(X,X') | Y\smash{=}Y' \right] + l_0 \mathbb{E} \left[ k(X,X') \right] \\
    = \frac{\Delta l}{M} \ \mathbb{E}_{X,X' \sim Pos} \left[ k(X,X') \right] + l_0 \mathbb{E} \left[ k(X,X') \right],
\end{dmath}
as $ \mathbb{P}(Y\smash{=}Y') = 1 / M$ and $ \mathbb{E}_{X,X'} \left[ k(X,X') | Y\smash{=}Y \right] = \mathbb{E}_{X,X' \sim Pos} \left[ k(X,X') \right] $. The second one, using the independence between $X'$ and $Y''$, yields:
\begin{dmath}
    \mathbb{E} \left[ k(X,X') l(Y,Y'') \right] = \mathbb{E}_{X,Y} \left[ \mathbb{E}_{X'} \left[ k(X,X') \left( \Delta l \ \underbrace{\mathbb{E}_{Y''} \left[ \one_{\{ Y = Y'' \} } \right]}_{1/M} + l_0  \right) \right] \right] \\
    = \left( \frac{\Delta l}{M} + l_0  \right) \mathbb{E}_{X,Y} \left[ \mathbb{E}_{X'} \left[ k(X,X') \right] \right] \\
    = \left( \frac{\Delta l}{M} + l_0  \right) \mathbb{E} \left[ k(X,X') \right].
\end{dmath}
Finally, as $ \mathbb{E} \left[ l(Y,Y'') \right] = \frac{\Delta l}{M} + l_0 $, the last term is equal to the second one, and it gets cancelled. We thus have
\begin{dmath}
    \text{HSIC}(X,Y) = \frac{\Delta l}{M} \mathbb{E}_{X,X' \sim Pos} \left[ k(X,X') \right] + l_0 \mathbb{E} \left[ k(X,X') \right] - \left( \frac{\Delta l}{M} + l_0  \right) \mathbb{E} \left[ k(X,X') \right] \\
    = \frac{\Delta l}{M} \mathbb{E}_{X,X' \sim Pos} \left[ k(X,X') \right] - \frac{\Delta l}{M} \mathbb{E} \left[ k(X,X') \right].
\end{dmath}

\end{proof}


\end{document}